\pdfoutput=1

\documentclass[11pt]{article}

\usepackage[preprint]{ACL2023}

\usepackage{times}
\usepackage{latexsym}

\usepackage[T1]{fontenc}

\usepackage[utf8]{inputenc}

\usepackage{microtype}

\usepackage{inconsolata}

\newcommand\methodname{\textcolor{black}{\textsc{ProxyLM}}}

\usepackage{amsmath}
\usepackage{multirow}
\usepackage{bbm}
\usepackage{graphicx}
\usepackage{amssymb}
\usepackage{booktabs}
\usepackage{lscape}
\usepackage{pifont}
\usepackage{algorithm}
\usepackage{graphicx}
\usepackage{caption}
\usepackage{subcaption}
\usepackage{float}

\usepackage{xcolor}
\definecolor{blue}{RGB}{53, 130, 230}

\newcommand{\cmark}{\ding{51}}%
\newcommand{\xmark}{\ding{55}}%

\title{ProxyLM: Predicting Language Model Performance on Multilingual \\Tasks via Proxy Models}

\author{David Anugraha$^{1}$, Genta Indra Winata$^{2}$\thanks{\text{ }The work was conducted outside Capital One.}, Chenyue Li$^1$, \\
\textbf{Patrick Amadeus Irawan$^{3}$}, \textbf{En-Shiun Annie Lee$^{1,4}$}\\
  $^1$University of Toronto \quad $^2$Capital One \\
  $^3$Institut Teknologi Bandung \quad $^4$Ontario Tech University\\
  \texttt{anugraha@cs.toronto.edu, genta.winata@capitalone.com,} \\
  \texttt{\{jimschenchen,patrickamadeusirawan\}@gmail.com, annie.lee@ontariotechu.ca}}

\begin{document}
\maketitle
\begin{abstract}
Performance prediction is a method to estimate the performance of Language Models (LMs) on various Natural Language Processing (NLP) tasks, mitigating computational costs associated with model capacity and data for fine-tuning. Our paper presents \methodname{}, a scalable task- and language-agnostic framework designed to predict the performance of LMs using proxy models. These proxy models act as surrogates, approximating the performance of the LM of interest. By leveraging these proxy models, \methodname{} significantly reduces computational overhead in task evaluations, achieving up to a 37.08$\times$ speedup over traditional methods, even with our smallest proxy models. Our results across multiple multilingual NLP tasks and various robustness tests demonstrate that \methodname{} not only adapts well to previously unseen languages in pre-trained LMs, but also generalizes effectively across different datasets, outperforming the state-of-the-art by at least 1.78$\times$ in terms of root-mean-square error (RMSE).
\end{abstract}

\section{Introduction}

Language Models (LMs) have become increasingly valuable for assessing Natural Language Processing (NLP) tasks~\cite{raffel2020exploring, brown2020language, costa2022no, touvron2023llama1, touvron2023llama, le2023bloom}. However, fine-tuning and evaluating these models are resource-intensive processes in terms of both computation and time. These costs escalate with model size, especially when experimenting across multiple datasets. As highlighted in \citet{kaplan2020scaling}, there is a scaling law that applies to both model and dataset sizes, and computational demands, indicating that larger models and broader datasets require increased computational resources. Modeling low-resource languages (LRLs) in multilingual contexts presents a range of challenges. One significant challenge is the limited data availability, which hampers effective fine-tuning processes~\cite{gu2018universal,adilazuarda2024lingualchemy}, making model adaptation through fine-tuning a challenging task~\cite{zoph2016transfer,liu2021continual}. Another critical issue is the lack of pre-training data for numerous regional languages, such as Southeast Asian languages~\cite{winata2022cross,winata2024miners,yong2024lexc}, with many languages being omitted during the pre-training phase of multilingual LMs.

\begin{figure*}[!th]
    \centering
    \includegraphics[width=0.97\linewidth]{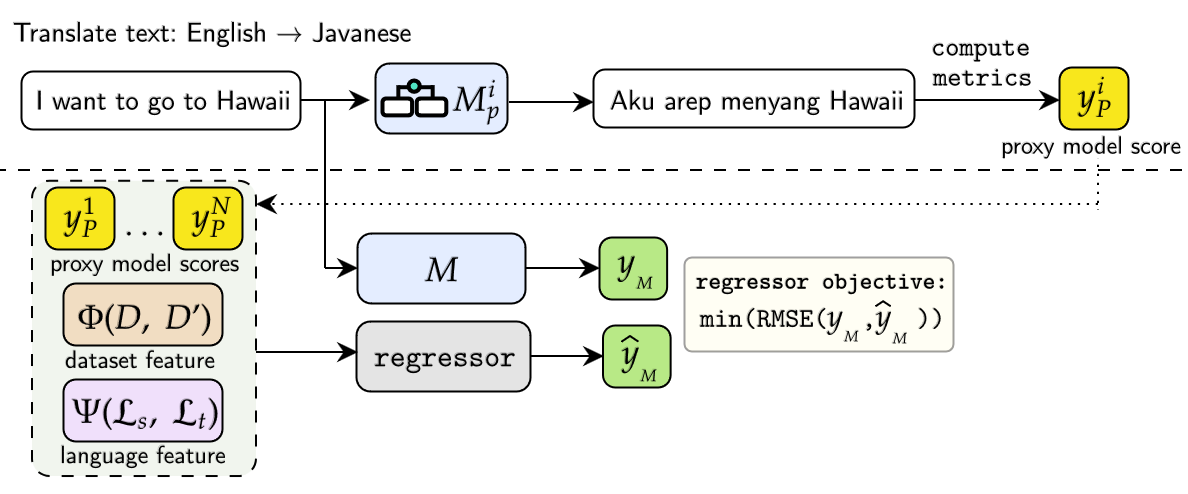} 
    \caption{\methodname{} framework for LM performance prediction. \textbf{(Top)} The evaluation metric is computed on the test set using \textcolor{blue}{\textbf{a proxy model}} $\mathcal{M}_p^i$. \textbf{(Bottom)} The regressor $g$ is trained using proxy model scores as well as dataset and language features by minimizing the RMSE difference of $y_\mathcal{M}$ and $\hat{y}_\mathcal{M}$.}
    \label{fig:pipeline}
    \vspace{-1mm}
\end{figure*}

Given the limited academic computational resources for LM fine-tuning and inadequate datasets for LRLs, performance prediction is an efficient method to alleviate the dependency on extensive resources by leveraging past performance records on an NLP task. While linear regression and gradient-boosting hold promise in performance prediction~\cite{birch2008predicting,srinivasan2021predicting,xia2020predicting,ye2021towards,schram2023performance,khiu2024predicting}, existing solutions primarily focus on homogeneous data settings and prioritize high-resource languages using Transformer models~\cite{vaswani2017attention}.
\citet{khiu2024predicting} examine diverse datasets and LRLs but encounter limitations in the number of experiments, language diversity, and model scope, focusing solely on mBART~\cite{liu2020multilingual}. Recent advancements in larger multilingual models, like NLLB~\cite{costa2022no} and M2M100~\cite{fan2021beyond}, have significantly improved machine translation capabilities, exceeding those of mBART and other LMs~\cite{zhu2023multilingual}.

In this paper, we propose \methodname{},\footnote{We release our code at~\url{https://github.com/davidanugraha/proxylm}.} 
a scalable task- and language- agnostic framework to predict LM performance by utilizing proxy models. Proxy models are defined as substitute models, wherein the performance of these substitute models are used to estimate the performance of another LM. This other model can be significantly larger than our proxy models. For optimizing the prediction, we utilize much smaller LMs as proxy models and off-the-shelf models without further tuning. This approach is very scalable to multiple proxy models and task-agnostic to any modalities, thus it can be applied to any downstream tasks. This study focuses on three multilingual tasks, covering machine translation (MT), intent classification, and slot filling. Our approach outperforms the existing work from \citet{xia2020predicting,ye2021towards,schram2023performance,khiu2024predicting}, which opens a new avenue to employ LMs for model performance prediction. Therefore the contribution of our paper can be summarized in three-fold:
\begin{enumerate}
    \item We introduce \methodname{}, an efficient and scalable task- and language-agnostic framework designed to predict the performance of LMs. This framework significantly reduces the computational costs associated with fine-tuning and inference during model selection.
    \item We demonstrate the effectiveness and robustness of \methodname{} across 34 dataset sources and 56 languages for MT, and 51 languages for intent classification and slot filling, each on two estimated LMs. Our framework substantially outperforms all existing baselines in all NLP tasks, datasets, and settings, including scenarios involving extremely LRLs that remain unseen by pre-trained LMs and across different datasets, surpassing the state-of-the-art performance measured with root-mean-square error (RMSE) by at least 1.78$\times$.
    \item We also provide a time analysis comparing the deployment of proxy models for performance prediction to direct LM fine-tuning. Our results indicate that, with our smallest proxy models, we can achieve up to a 37.08$\times$ speedup on task evaluation compared to the traditional approach, highlighting the efficiency of our approach.
\end{enumerate}

\section{Methodology}

In this section, we formally define the LM performance prediction problem and our proposal to improve performance prediction.

\subsection{\methodname{}}
Recall that performance prediction is a task of estimating a system's performance based on the model and its training strategy, training and test dataset, and language used. Formally, let LM $\mathcal{M}$ be our estimated model. $\mathcal{M}$ is trained over a training dataset $\mathcal{D}$ with source language $\mathcal{L}_s$ and target language $\mathcal{L}_t$, and then tested using dataset $\mathcal{D}'$. $\mathcal{M}$'s performance, denoted $y_{\mathcal{M}}$, can be formulated under function $f$ that relates between these variables:
\begin{equation}
    y_{\mathcal{M}} = f(\mathcal{M}, \mathcal{D}, \mathcal{D}', \mathcal{L}_s, \mathcal{L}_t).
\label{eq:original_perfpred}
\end{equation}
We can approximate $f$ by transforming Equation $\ref{eq:original_perfpred}$ into a regression task with a regressor function $g$, which will be trained on past performance records. Previous works ~\cite{xia2020predicting,ye2021towards,schram2023performance,khiu2024predicting} formulate regressor that takes dataset features $\Phi(\mathcal{D}, \mathcal{D}')$ to identify the characteristics of the training and test datasets, as well as the distribution shift between them. It also takes language features $\Psi(\mathcal{L}_s, \mathcal{L}_t)$ to measure the similarities between the source and target languages.
This can be formulated as follows:
\begin{equation}
\hat{y_{\mathcal{M}}} = g(\Phi(\mathcal{D}, \mathcal{D}'); \Psi(\mathcal{L}_s, \mathcal{L}_t)).
\label{eq:approx_perfpred}
\end{equation}
We present \methodname{}, a framework that leverages the past performance of other models, referred to as proxy models, as additional context for our regressor. Intuitively, proxy models can provide valuable insights that assist in predicting the performance of the estimated model $\mathcal{M}$, which addresses the gap in previous works for not accounting $\mathcal{M}$. Formally, let $\mathcal{M}_p = [\mathcal{M}_p^1, \dots, \mathcal{M}_p^N]$ be a set of $N$ proxy models. To integrate the information from these proxy models, we modify Equation \ref{eq:approx_perfpred} as follows:
\begin{equation}
\hat{y_{\mathcal{M}}} = g(\hat{y}_{\mathcal{M}_p};\Phi(\mathcal{D}, \mathcal{D}'); \Psi(\mathcal{L}_s, \mathcal{L}_t)),
\label{eq:approx_perfpred_proxylm}
\end{equation}
where $y_{\mathcal{M}_p} = [y_{\mathcal{M}_p^1}, \dots, y_{\mathcal{M}_p^N}]$ represents the  performance records of $N$ proxy models.
The advantage of using proxy models arises from their faster fine-tuning and evaluation compared to the estimated model $\mathcal{M}$. This also means that off-the-shelf models can be used directly without additional tuning if they already perform the task adequately, further enhancing efficiency.

\subsection{\methodname{} Features}

\paragraph{Language Features.}
We use URIEL Typological Database~\cite{littell2017uriel} similar to~\citet{xia2020predicting} including geographic, genetic, inventory, syntactic, phonological, and featural distance. The language features are useful to provide a language-specific representation to the regressor.

\paragraph{Dataset Features.} We extract six features from the dataset, including train size, vocab size, average sentence length, word overlap, Type-Token Ratio (TTR), and TTR distance from $\mathcal{D}$ and $\mathcal{D'}$ based on ~\citet{xia2020predicting}. We will refer to these features and language features combined as NLPerf features. Furthermore, we incorporate the distribution shift information between the training and test datasets using Jensen-Shannon Divergence (JSD) as described by~\citet{khiu2024predicting}. In addition, we include the cosine similarity of term frequency-inverse document frequency (TF-IDF) representations and sentence embeddings using Sentence-BERT~\cite{reimers2019sentencebert}. Details on how these features are computed can be found in Appendix Section \ref{sec:app-detail-dataset-feat}.

\paragraph{Proxy Models Features.} 
We leverage the performance data from proxy models, derived by averaging results from multiple fine-tuning and evaluation iterations on identical datasets and languages. Moreover, we retain the flexibility to adjust the number of proxy models employed, facilitating efficient and adaptable performance estimation.

\section{Experimental Setup}

In this section, we describe the datasets and LMs used to obtain LMs' performance records. These records are then used to train various regressor models under different experimental settings to investigate our approach to performance predictions. The details of the hyper-parameters for both the LMs and the regressors are provided in \ref{hyperparam}.

\subsection{Datasets}
\paragraph{Machine Translation.} We use two types of datasets: English-centric and Many-to-Many Languages. The English-centric dataset involves English serving as either the source or target language. Our English-centric dataset comes from the MT560~\cite{gowda2021many} dataset, where we curate 32 datasets and select 50 languages out of 500 for evaluation. Furthermore, we use the FLoRes-200 dataset~\cite{costa2022no} for additional validation and test sets. These datasets consist of translations with varying quality across diverse domains, presenting significant challenges for a robust performance prediction. In contrast, the Many-to-Many Languages dataset allows any language to act as the source or target. We use the NusaTranslation dataset~\cite{cahyawijaya2023nusawrites} as our Many-to-Many Languages dataset, which comprises parallel texts in 12 Indonesian regional languages. As many of these languages are absent in pre-trained multilingual models, we analyze 8 out of the 12 languages due to limited data in the remaining 4. Both datasets encompass 56 languages across various domains such as economics, technology, and medicine. Detailed language insights are available in the Appendix Section \ref{sec:more-info-langs}.

\paragraph{Intent Classification and Slot Filling.}
We use MASSIVE \cite{fitzgerald2022massive} as our dataset encompassing 51 languages. We fine-tune each LM on all languages and evaluate it on all languages. Detailed language insights are available in the Appendix Section \ref{sec:more-info-langs}.

\subsection{Estimated LMs}

\paragraph{Machine Translation.} We employ two estimated LMs: M2M100 1.2B~\cite{fan2021beyond} and NLLB 1.3B~\cite{costa2022no}. Each estimated model is fine-tuned using a standard next-token prediction objective on the training set.

\paragraph{Intent Classification and Slot Filling.} We employ two decoder-only estimated LMs: LLaMA-3 Instruct (8B) ~\cite{dubey2024llama} and Aya-23 (8B)~\cite{aryabumi2024aya}. Both models are fine-tuned using supervised fine-tuning (SFT) combined with Low-Rank Adaptation (LoRA)~\cite{hu2021lora} on the training set.

\subsection{Proxy Models}

\paragraph{Machine Translation.} We utilize four different transformer-based models: an encoder-decoder random initialized Transformers (100M)~\cite{vaswani2017attention}, SMaLL-100 (330M)~\cite{mohammadshahi2022small}, M2M100~\cite{fan2021beyond}, and NLLB ~\cite{costa2022no}. For M2M100 and NLLB, we use the models without any additional tuning \textbf{(No FT)} in a zero-shot fashion. For simplicity, the term ``fine-tuning" will be used throughout this paper to refer to both the process of training from scratch (as in the case of the Transformer (100M) model) and the process of fine-tuning pre-trained LMs. Model details are provided in the Appendix Section ~\ref{paragraph-models-details}. The evaluation is primarily conducted using SentencePiece BLEU (spBLEU)~\cite{goyal2022flores}, which has proven to be a reliable metric in multilingual and LRLs. For a more comprehensive assessment, we also use COMET-22 \cite{rei2022comet}, as it shows a high correlation with human judgments \cite{freitag2023results}. However, COMET-22 is applied to only a subset of the English-centric dataset due to its limited language coverage, especially to LRLs.

\paragraph{Intent Classification and Slot Filling.} We utilize three decoder-only models: SmolLM (135M and 360M) \cite{allal2024SmolLM} and BLOOMZ (560M) \cite{muennighoff2022crosslingual}. Model details are provided in the Appendix Section ~\ref{paragraph-models-details}. The evaluation is done using accuracy for intent classification and micro-F1 for slot filling.

\subsection{Regressor Models} 

We utilize XGBoost~\cite{chen2016xgboost}, LGBM~\cite{ye2021towards}, Poly2~\cite{khiu2024predicting}, and Poly3~\cite{khiu2024predicting} as our regressors. In most of our experiments, we apply XGBoost as our default regressor because we find it to be the best-performing model based on the cross-validation during training, while the other regressors serve as baselines. Specifically for MT in the Many-to-Many Languages setting, Matrix Factorization with context features (MF) is used as an additional baseline~\cite{schram2023performance}. We do not apply MF to our English-centric setting because MF requires the performance records to be structured in two dimensions—one for the source language and one for the target language. In the English-centric setting, this would result in a sparse matrix with only one fully populated row or column, corresponding to English, making MF impractical for this setup.

\begin{table*}[!th]
\centering
\resizebox{\textwidth}{!}{
    \begin{tabular}{lrrrrr|rrrrr}
    \toprule
     & \multicolumn{4}{c}{\textbf{English-centric}} & & \multicolumn{4}{c}{\textbf{Many-to-Many}} & \\
    \textbf{Models} & \multicolumn{2}{c}{\textbf{Random}} & \multicolumn{2}{c}{\textbf{LOLO}} & \multicolumn{1}{c|}{\textbf{Avg.}} & \multicolumn{2}{c}{\textbf{Random}} & \multicolumn{2}{c}{\textbf{LOLO}} & \multicolumn{1}{c}{\textbf{Avg.}} \\
    & \multicolumn{1}{c}{\textbf{M2M100}$\downarrow$} & \multicolumn{1}{c}{\textbf{NLLB}$\downarrow$} & \multicolumn{1}{c}{\textbf{M2M100}$\downarrow$} & \multicolumn{1}{c}{\textbf{NLLB}$\downarrow$} & &{\textbf{M2M100}$\downarrow$} & \multicolumn{1}{c}{\textbf{NLLB}$\downarrow$} & \multicolumn{1}{c}{\textbf{M2M100}$\downarrow$} & \multicolumn{1}{c}{\textbf{NLLB}$\downarrow$} & \\
    \midrule 
    \multicolumn{5}{l}{NLPerf~\cite{xia2020predicting} with different regressors} & \\ \midrule
    XGBoost & 7.69 & 7.73 & 9.20 & 12.92 & 9.39 & 2.45 & \textbf{1.11} & 7.83 & 8.28 & 4.94 \\
    Poly2~\cite{khiu2024predicting} & 11.21 & 16.23 & 15.55 & 43.02 & 21.50 & 4.70 & 4.68 & 7.07 & 7.90 & 6.09\\
    Poly3~\cite{khiu2024predicting} & 11.00 & 15.64 & 62.29 & 236.29 & 81.31 & 4.60 & 4.64 & 7.26 & 8.01 & 6.13 \\
    LGBM~\cite{ye2021towards} & 7.88 & 8.15 & 9.71& 12.81 & 9.64 & 3.65 & 2.60 & 7.08 & 7.14 & 5.12 \\
    \midrule
    \multicolumn{5}{l}{\methodname{} (Ours)$^{\ddagger}$ with different proxy models} & \\ \midrule
    Transformer & 4.68 & 7.22 & 6.18 & 11.78 & 7.47 & 2.56 & 1.70 & 5.65 & 6.24 & 4.04 \\
    SMaLL-100 & \underline{4.07} & 6.33 & \underline{4.59} & 10.33 & 6.33 & 2.56 & 1.65 & \underline{4.85} & \underline{5.14} & \underline{3.55} \\
    SMaLL-100 (No FT) & 5.27 & 6.04 & 6.28 & 10.94 & 7.13 & 2.44 & 1.34 & 6.93 & 7.25 & 4.49 \\
    Estimated Model (No FT) & 5.23 & \underline{4.15} & 6.18 & \underline{5.42} & \underline{5.25} & \textbf{2.38} & \underline{1.27} & 5.10 & 5.50 & 3.56 \\
    Ensemble$^\dagger$ & \textbf{3.21} & \textbf{3.68} & \textbf{3.74} & \textbf{4.94} & \textbf{3.89} & \underline{2.41} & 1.56 & \textbf{3.73} & \textbf{3.79} & \textbf{2.90}
    \\ \bottomrule
    \end{tabular}
}
\caption{MT test results on English-centric and Many-to-Many Languages datasets using spBLEU in average RMSE (\textbf{lower is better)}. \textbf{Bold} numbers indicate the best performance, while \underline{underlined} numbers represent the second-best performance. The columns show the setting and estimated model. ``No FT" denotes ``no fine-tuning" and the model inference is done in a zero-shot fashion. Avg represents the average of the results across the row for each respective dataset. $^{\ddagger}$The reported results use XGBoost as the regressor. $^\dagger$Ensemble denotes combining all four proxy models, the detailed breakdown of this result with the standard deviation can be seen in the Appendix Section~\ref{sec:ensemble-breakdown}.}
\label{tab:results-mt}
\end{table*}

\subsection{Experimental Settings}

Each regressor is evaluated using RMSE as our performance metric and evaluated 5 times. For all tasks, we set our experiment settings as follows:
\begin{itemize}
    \item \textbf{Random}: We randomly sample the performance records into training and test sets with a ratio of 7:3. Then, we run 10-fold cross-validation on the training set to find the best hyper-parameters for each regressor. The best-performing regressor would subsequently be evaluated on the test set.
    \item \textbf{Leave-One-Language-Out (LOLO)}: We select one language as the test set, which is not encountered during training. 
\end{itemize}

To test the robustness of \methodname{}, we also provide two additional setups specifically for MT:
\begin{itemize}
    \item \textbf{Unseen}: The performance records for MT can be divided into two categories: (1) records with ``seen" languages and (2) records with ``unseen" languages. ``Unseen" languages refer to languages that are not present in the pre-training LM data, while ``seen" languages denote those that are present. In this setting, the regressor is trained using records of ``seen" languages and tested using records of ``unseen" languages.
    \item \textbf{Cross-Dataset}: The regressor can be trained using performance records from the English-centric dataset and tested using the Many-to-Many Languages dataset. We opt not to reverse this setup as the Many-to-Many dataset exhibits no domain shift and contains fewer performance records.
\end{itemize}

\section{Results and Analysis}

In this section, we present the results of the performance predictions for \methodname{} and baselines under the specified settings. Further, we discuss the robustness, effectiveness, and efficiency of \methodname{} in the context of performance prediction.

\subsection{Machine Translation}

\subsubsection{English-centric Results}
Table~\ref{tab:results-mt} shows the overall results on the English-centric dataset using spBLEU. $\methodname{}$ remarkably outperforms all existing baselines. We find that incorporating all proxy models (Ensemble) is the most effective for prediction, leading to a 2.41$\times$ averaged reduction in RMSE across all experimental settings compared to the best baseline. Note that the significant improvement remains consistent when evaluated using a different metric, such as COMET-22, which yields a 2.00× averaged reduction in RMSE across experimental settings compared to the best baseline, as shown in Table \ref{tab:results-english-centric-comet} in Appendix Section \ref{sec:app-detail-dataset-feat}. We observe that using the \textbf{No FT} estimated model to predict the performance of their fine-tuned models is surprisingly useful in all settings, especially for NLLB, where the model already has decent machine translation quality on LRLs. This observation is supported by our findings within the XGBoost model that the NLLB No FT feature has the highest importance score among all features, as shown in Figure \ref{fig:nllb-mt560-feat} in the Appendix. Furthermore, using SMaLL-100 fine-tuned performance provides useful estimations for settings involving M2M100 as the estimated model. This may indicate that the performance of a model with similar architecture can be a good estimator for the performance of the larger estimated model. In other words, the choice of proxy model to help prediction matters. Feature importance analysis from the XGBoost model supports this, revealing that the SMaLL-100 fine-tuned feature has the highest importance score among all features, as shown in Figure \ref{fig:m2m100-mt560-feat} in the Appendix.



\begin{table*}[!th]
\centering
\resizebox{0.98\textwidth}{!}{
    \begin{tabular}{lccccc|ccccc}
    \toprule   
    & \multicolumn{4}{c}{\textbf{Intent Classification}} &  & \multicolumn{4}{|c}{\textbf{Slot Filling}} & \\
    \textbf{Models} & \multicolumn{2}{c}{\textbf{Random}} & \multicolumn{2}{c}{\textbf{LOLO}} & \textbf{Avg.} & \multicolumn{2}{c}{\textbf{Random}} & \multicolumn{2}{c}{\textbf{LOLO}} & \textbf{Avg.} \\
    & \multicolumn{1}{c}{\textbf{Aya23}$\downarrow$} & \multicolumn{1}{c}{\textbf{LLaMA3}$\downarrow$} & \multicolumn{1}{c}{\textbf{Aya23}$\downarrow$} & \multicolumn{1}{c}{\textbf{LLaMA3}$\downarrow$} &  & \multicolumn{1}{c}{\textbf{Aya23}$\downarrow$} & \multicolumn{1}{c}{\textbf{LLaMA3}$\downarrow$} & \multicolumn{1}{c}{\textbf{Aya23}$\downarrow$} & \multicolumn{1}{c}{\textbf{LLaMA3}$\downarrow$} &  \\
    \midrule 
    \multicolumn{5}{l}{NLPerf~\cite{xia2020predicting} with different regressors} & \\ \midrule
    XGBoost &
    0.0761 & 0.0191 & 0.1573 & 0.0581 & 0.0777 & 0.0693 & 0.0548 & 0.1219 & 0.1093 & 0.0888 \\
    Poly2~\cite{khiu2024predicting} & 0.1996 & 0.0979 & 0.2075 & 0.0918 & 0.1492 & 0.1396 & 0.1412 & 0.1418 & 0.1414 & 0.1410 \\
    Poly3~\cite{khiu2024predicting} & 0.1990 & 0.0969 & 0.2191 & 0.0925 & 0.1519 & 0.1393 & 0.1401 & 0.1448 & 0.1413 & 0.1414 \\
    LGBM~\cite{ye2021towards} & 0.0839 & 0.0198 & 0.1545  & 0.0558 & 0.0785 & 0.0692 & 0.0557 & 0.1218 & 0.1152 & 0.0905 \\
    \midrule
    \multicolumn{5}{l}{\methodname{} (Ours)$^{\ddagger}$ with different proxy models} & \\ \midrule
    SmolLM (135M) & 0.0676 & 0.0171 & 0.1273 & 0.0455 & 0.0644 & 0.0618 & 0.0538 & 0.1004 & 0.0953 & 0.0778 \\
    SmolLM (360M) & \textbf{0.0604} & \textbf{0.0157} & \underline{0.1118} & \textbf{0.0441} & \textbf{0.0580} & \textbf{0.0562} & \textbf{0.0506} & \underline{0.0844} & \textbf{0.0868} & \textbf{0.0695} \\
    BLOOMZ (560M) & 0.0692 & 0.0179 & 0.1283 & 0.0482 & 0.0659 & 0.0618 & 0.0540 & 0.1023 & 0.0995 & 0.0794 \\
    Ensemble$^\dagger$ & \underline{0.0609} & \underline{0.0164} & \textbf{0.1112} & \underline{0.0442} & \underline{0.0582} & \underline{0.0561} & \underline{0.0508} & \textbf{0.0830} & \underline{0.0884} & \underline{0.0696}
    \\ \bottomrule
    \end{tabular}
}
\caption{Intent classification and slot filling results using accuracy and micro-F1 score, respectively, with average RMSE (\textbf{lower is better}). \textbf{Bold} numbers indicate the best performance, while \underline{underlined} numbers represent the second-best performance. Avg represents the average of the results across the row for each respective task. $^{\ddagger}$The reported results use XGBoost as the regressor for both intent classification and slot filling. $^\dagger$Ensemble denotes combining all three proxy models. The detailed breakdown of this result with the standard deviation can be seen in the Appendix Section~\ref{sec:ensemble-breakdown}.}
\label{tab:results-intent-slot-combined}
\end{table*}

\subsubsection{Many-to-Many Languages Results}
Table~\ref{tab:results-mt} presents the performance of different models on the Many-to-Many Languages dataset. The results reveal that the Ensemble model achieves the lowest RMSE, with a 1.70$\times$ averaged reduction in RMSE across all experimental settings compared to the best baseline, indicating superior accuracy in performance predictions. An exception occurs in the random NLLB setting, where the model utilizing only NLPerf features outperforms the ensemble model, achieving the best performance. Note that no domain shift occurs within the dataset.

A comparative analysis shows that predicting the performance of the M2M100 model in the random setting presents a greater challenge compared to predicting the NLLB model. This discrepancy suggests that the complexity of performance prediction can vary substantially depending on the specific LM and the conditions under which it is evaluated. A particularly noteworthy finding is the effectiveness of using No FT models for estimating LM performance. The No FT models, which do not require any additional fine-tuning, demonstrate high accuracy in their performance predictions. This method offers substantial efficiency benefits, as it eliminates the need for extensive computational resources typically required for model training. In contrast, we find similar results between the LOLO setting for Many-to-Many Languages and English-centric results, where $\methodname{}$ using Ensemble remarkably outperforms all existing baselines. In addition, we find that using SMaLL-100 fine-tuned performance results in better predictions compared to those of the No FT estimated model.


\subsection{Intent Classification and Slot Filling}

Table~\ref{tab:results-intent-slot-combined} presents the overall results for both intent classification and slot filling tasks. In these experiments, \methodname{} employs XGBoost as the regressor for intent classification and LGBM for slot filling, as LGBM exhibited superior performance during training cross-validation for the slot filling task. Consistent with our findings in MT, \methodname{} outperforms all existing baselines in both tasks. Notably, the SmolLM (360M) model emerges as the most effective proxy model, achieving an average RMSE reduction of 1.34$\times$ in intent classification and 1.28$\times$ in slot filling across all experimental settings when compared to the best baseline. As with MT, the choice of proxy model significantly impacts prediction performance. Based on the feature importance analysis in Figure \ref{fig:slot-aya-feat}-\ref{fig:intent-llama-feat} in the Appendix, SmolLM (360M) exhibits the highest importance score among all proxy models in both tasks. This suggests that the size of the proxy models may not be a reliable indicator of their effectiveness.

\section{Ablation Study}

\subsection{Robustness of \methodname{}}

\begin{figure}[!t]
    \includegraphics[width=\columnwidth]{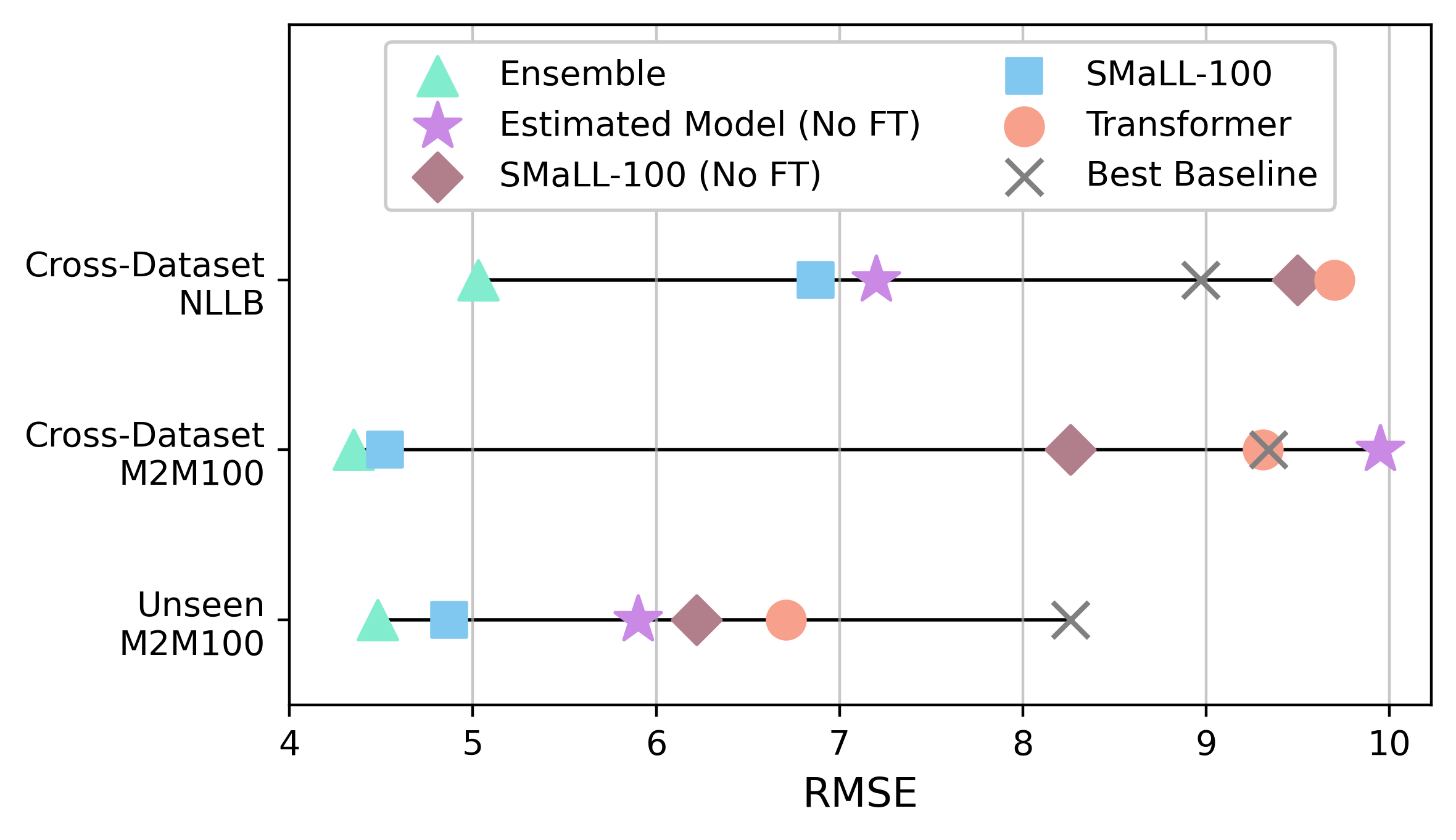}
    \caption{Unseen and Cross-Dataset MT test results on English-centric dataset in average RMSE (\textbf{lower is better)}. We only show the best-performing baseline for comparison with \methodname{} with different proxy models. ``No FT" denotes ``no fine-tuning". We only show M2M100 results for the Unseen setting since NLLB covers all languages in the English-centric dataset. The reported results for the Unseen setting use XGBoost, while the Cross-Dataset experiments use LGBM. Ensemble denotes combining all four proxy models. The detailed breakdown of this result with the standard deviation can be seen in the Appendix Section~\ref{sec:ensemble-breakdown}.}
\label{fig:result-robust}
\end{figure}

Figure~\ref{fig:result-robust} illustrates the performance of our models in both the Unseen and Cross-Dataset setups, highlighting the robustness results achieved by \methodname{}. For the Cross-Dataset evaluation, we opted for LGBM instead of XGBoost, as it demonstrated better performance during cross-validation on the training set. \methodname{} with Ensemble shows a significant reduction in RMSE compared to the best baseline: a 1.84$\times$ reduction in the Unseen setup, and reductions of 2.15$\times$ and 1.78$\times$ for M2M100 and NLLB in the Cross-Dataset scenario, respectively. This consistent performance across datasets and languages—including those not encountered during the regressor's training, such as unseen languages for the pre-trained LMs—emphasizes the model's generalization capabilities. We also observe better performance for M2M100 compared to NLLB, which may be attributed to NLLB's reliance on an English-centric dataset containing only seen languages, lacking examples of unseen languages for the regressor. This may indicate the importance of including instances of unseen languages in the regressor training dataset for achieving more robust predictions.

\begin{figure}[!t]
    \centering
        \begin{subfigure}[t]{0.49\textwidth}
        \centering
        \includegraphics[width=0.9\linewidth]{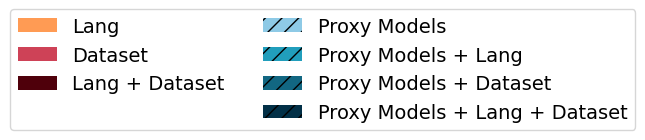} 
    \end{subfigure}
    \begin{subfigure}[t]{0.47\textwidth}
        \centering
        \includegraphics[width=\linewidth]{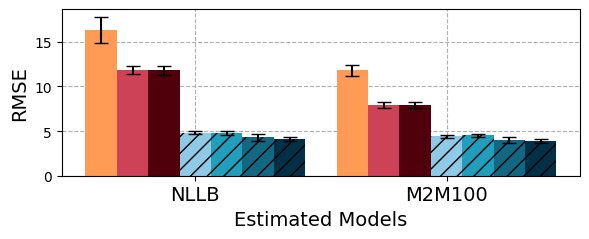} 
        \caption{English-centric test results.}
    \end{subfigure}
    \begin{subfigure}[t]{0.45\textwidth}
        \centering
        \includegraphics[width=\linewidth]{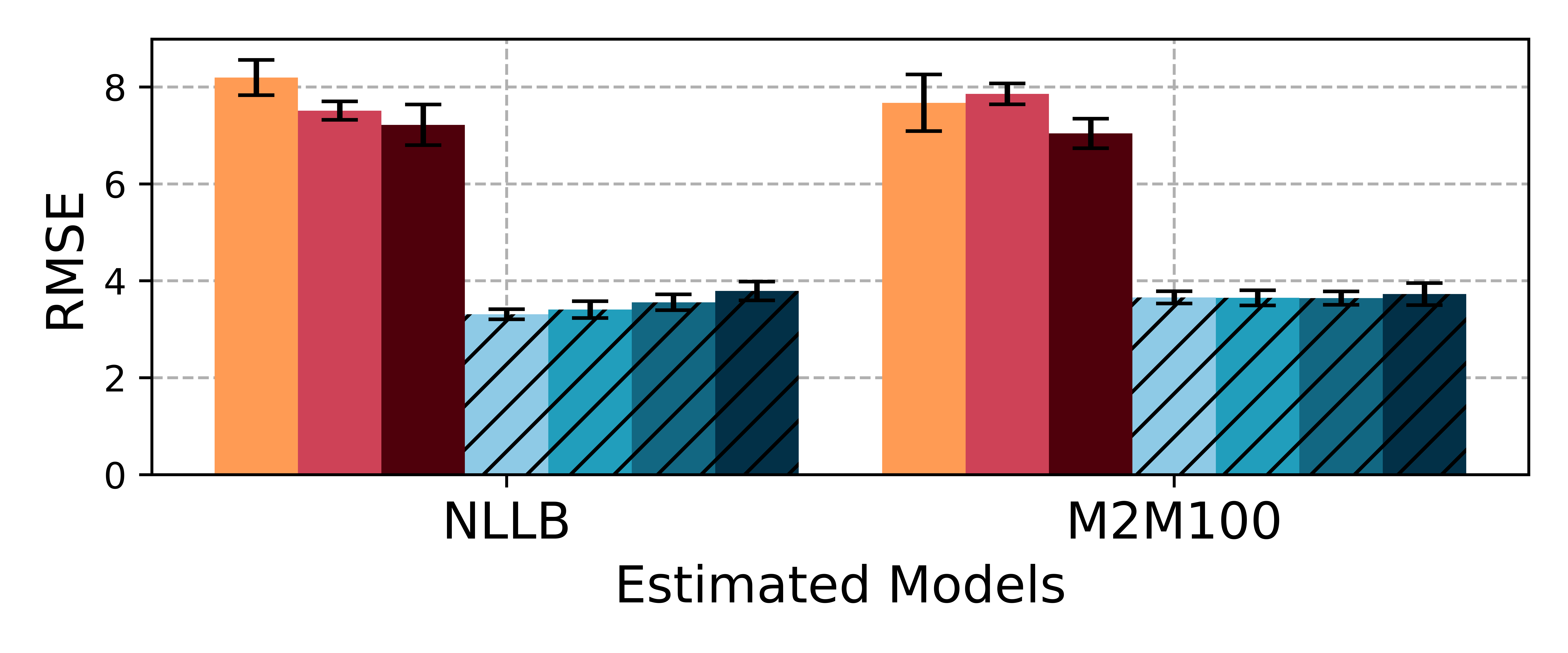} 
        \caption{Many-to-Many Languages test results.}
    \end{subfigure}
    \caption{Ablation study on the LOLO setting with XGBoost on English-centric and Many-to-Many Languages datasets. \textbf{Proxy Models} here indicates \textbf{Ensemble}, which is a combination of all proxy models. Proxy Models significantly reduce RMSE across all scenarios.}
\label{ablation-study}
\end{figure}

Figure~\ref{ablation-study} shows the impact of features used in \methodname{} for MT in the LOLO setting with XGBoost. Utilizing proxy models as features leads to a significant reduction in RMSE across all scenarios, showcasing their importance compared to other features. For the English-centric dataset, including language and dataset features alongside proxy models enhances performance. Dataset features alone show better improvement than language features alone, but the combination of both yields the best performance. On the other hand, for the Many-to-Many Languages dataset, the benefits of incorporating dataset and language features are less pronounced, especially for the M2M100, and there may even be a performance dip for the NLLB due to the dataset's lack of domain shift.

\begin{table}[!t]
\centering
\resizebox{.49\textwidth}{!}{
    \begin{tabular}{lrr@{ }l@{ }r@{ }l}
    \toprule   
    \textbf{Datasets} & \textbf{Inference} &\multicolumn{4}{c}{\textbf{Fine-tuning}} \\
    & & \multicolumn{2}{c}{\textbf{English-centric}} & \multicolumn{2}{c}{\textbf{Many-to-Many Langs.}} \\
    \midrule
    \textbf{Estimated models} \\
    \midrule
    M2M100 & 421 s & 3.94 & hrs (7.04$\times$) & 1.42 & hrs (7.32$\times$)  \\
    NLLB & 737 s & 12.08 & hrs (21.57$\times$) & 7.21 & hrs (37.08$\times$)  \\
    \midrule
    \textbf{Proxy models} \\
    \midrule
    SMaLL-100 & 333 s & 2.38 & hrs (4.25$\times$) & 1.03 & hrs (5.29$\times$)  \\
    Transformer & 231 s & 0.56 & hrs (1$\times$) & 0.19 & hrs (1$\times$) \\
    \bottomrule
    \end{tabular}
}
\caption{Comparison of LMs' inference time (in seconds) and fine-tuning time (in hours) for \textbf{one MT experimental run}. The multiplier of fine-tuning time is relative to the Transformer model. All times were calculated using the interquartile mean to ignore outliers.}
\label{tab:training-time-mt}
\end{table}

\begin{table}[!t]
\centering
\resizebox{.49\textwidth}{!}{
    \begin{tabular}{lr@{ }l@{ }r@{ }l}
    \toprule   
    \textbf{Datasets} & \multicolumn{2}{c}{\textbf{Intent Classification}} & \multicolumn{2}{c}{\textbf{Slot Filling}} \\
    \midrule
    \textbf{Estimated models} \\
    \midrule
    Aya-23 & 0.68 & hrs (6.13$\times$) & 5.10 & hrs (13.48$\times$)  \\
    LLaMA3 & 0.54 & hrs (4.91$\times$) & 4.36 & hrs (11.53$\times$)  \\
    \midrule
    \textbf{Proxy models} \\
    \midrule
    BLOOMZ-560M & 0.12 & hrs (1.06$\times$) & 0.59 & hrs (1.57$\times$)  \\
    SmolLM-360M & 0.12 & hrs (1.09$\times$) & 0.40 & hrs (1.05$\times$)  \\
    SmolLM-150M & 0.11 & hrs (1.00$\times$) & 0.38 & hrs (1.00$\times$) \\
    \bottomrule
    \end{tabular}
}
\caption{Comparison LMs' overall fine-tuning and evaluation time (in hours) for \textbf{one intent classification or slot filling experimental run}. The multiplier of the time is relative to SmolLM-135M model as it is the smallest proxy model. All times were calculated using the interquartile mean to ignore outliers.}
\label{tab:timing-finetune}
\end{table}

\begin{figure*}[!th]
    \centering
    \begin{subfigure}[t]{0.251\textwidth}
        \centering
        \caption{}
        \includegraphics[width=\linewidth]{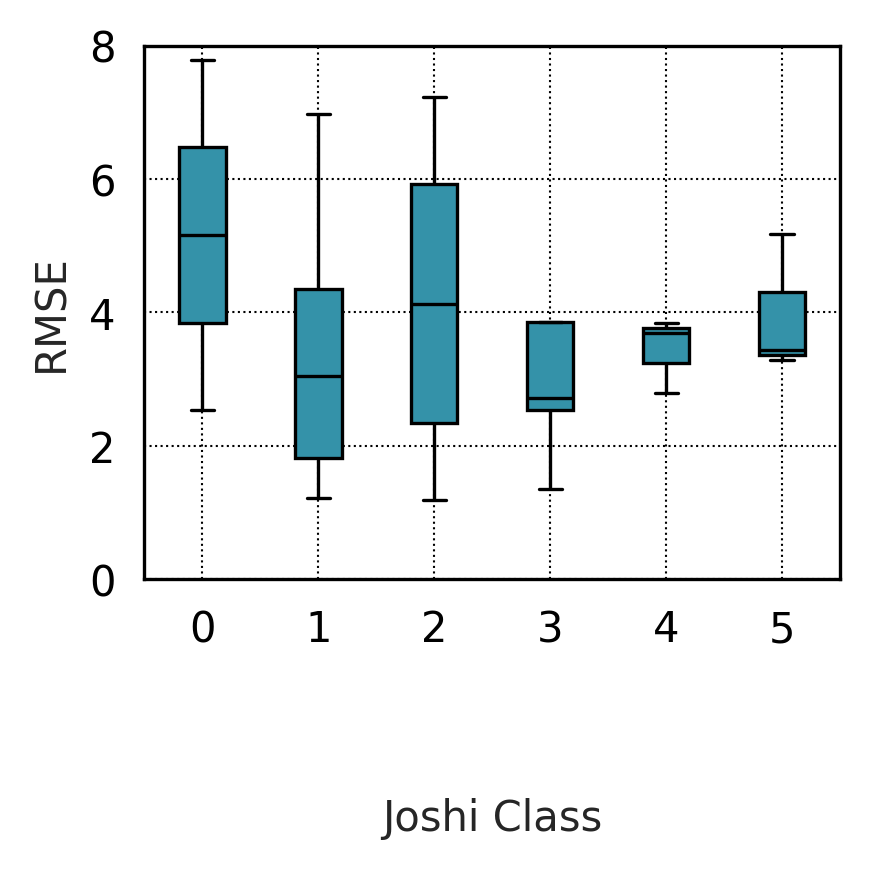} 
        
    \end{subfigure}
    \begin{subfigure}[t]{0.338\textwidth}
        \centering
        \caption{}
        \includegraphics[width=\linewidth]{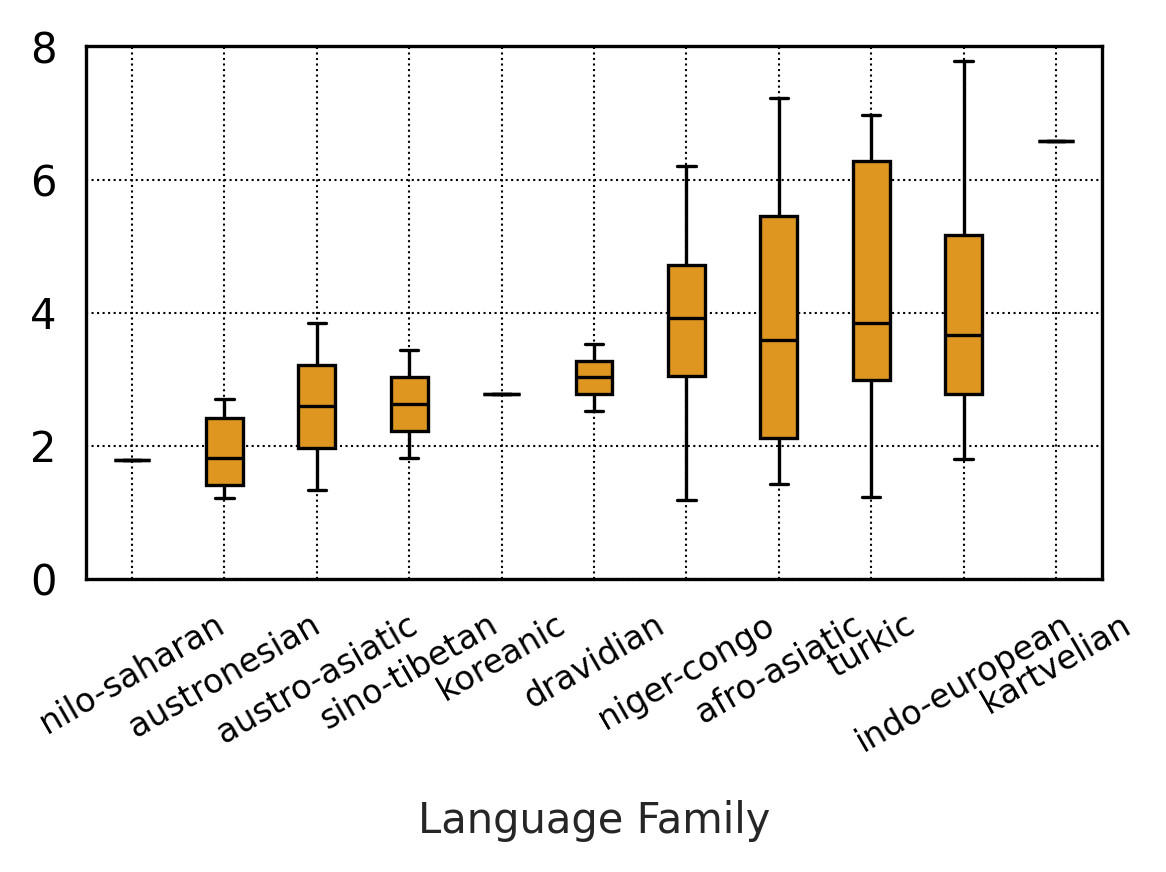} 
        
    \end{subfigure}
    \begin{subfigure}[t]{0.363\textwidth}
        \centering
        \caption{}
        \includegraphics[width=1\linewidth]{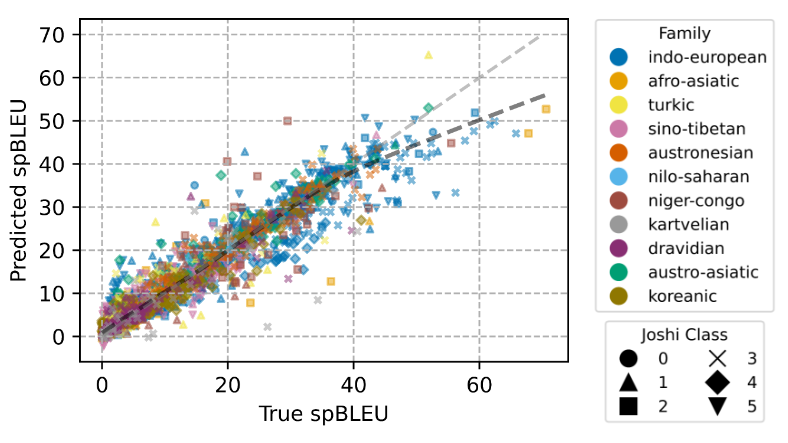} 
        
    \end{subfigure}
    \caption{Detailed results of XGBoost with \methodname{} Ensemble on \textbf{M2M100} model under the \textbf{LOLO} setting on MT using the English-centric dataset from Table \ref{tab:results-mt}. The results are grouped by (a) Joshi Class and (b) language family that follows the mapping which is provided in Appendix \ref{sec:more-info-langs}; (c) shows the scatter plot illustrating the correlation of spBLEU scores between the \methodname{}'s prediction and estimated LM, with the light gray dashed line representing the line of equality ($y = x$) with $R^2 = 0.90$ and black dashed line representing Locally Weighted Scatterplot Smoothing (LOWESS) curve to represent the trend.}
    \label{fig:lolo-m2m100-mt560-by-lang-group}
    \vspace{-0.5mm}
\end{figure*}

\subsection{Time Efficiency}
Table~\ref{tab:training-time-mt} compares the fine-tuning and inference times required for the estimated and proxy models on MT, while Table ~\ref{tab:timing-finetune} compares the fine-tuning and evaluation times required for the estimated and proxy models on intent classification and slot filling. The results demonstrate that fine-tuning proxy models or direct inference from any model is remarkably faster than fine-tuning all estimated models. Table~\ref{tab:timing-regressor} further illustrates this point, showing only a minimal trade-off in the time needed to train the regressor models. This additional training time is relatively negligible, highlighting the efficiency of using proxy models.

\begin{table}[!t]
\centering
\resizebox{.49\textwidth}{!}{
    \begin{tabular}{lrrr}
    \toprule   
    \textbf{Regressors} & \textbf{English-centric (MT)} & \textbf{Many-to-Many (MT)} & \textbf{Intent and Slot} \\
    \midrule
    XGBoost & 2.24 s & 0.87 s & 6.33 s  \\
    Poly2 & 0.07 s & 0.06 s & 0.09 s  \\
    Poly3 & 0.06 s & 0.06 s & 0.10 s \\
    LGBM & 90.79 s & 27.51 s & 118.24 s \\
    MF & N/A & 141.69 s & N/A \\
    \bottomrule
    \end{tabular}
}
\caption{Regressor models training time (in seconds) per one round of cross-validation with 10-folds across all setups. We combine the timing for intent classification and slot filling since they both contain the same amount of training data. All times were calculated using the interquartile mean to ignore outliers.}
\label{tab:timing-regressor}
\end{table}

\subsection{Performance by Language Categories}

In Figure \ref{fig:lolo-m2m100-mt560-by-lang-group}, we present detailed XGBoost results with \methodname{} Ensemble on the M2M100 model under the English-centric LOLO experiment, grouped by language categories. 
\methodname{} demonstrates relatively stable performance across languages belonging to different Joshi classes and linguistic families. Based on the Locally Weighted Scatterplot Smoothing (LOWESS) ~\cite{cleveland1979ze} curve depicted in Figure \ref{fig:lolo-m2m100-mt560-by-lang-group}(c), our method consistently maintains unbiased predictions for spBLEU scores below 40 across various language types. However, as the spBLEU score increases, the availability of data points diminishes, leading to our method under-predicting the performance compared to the true spBLEU score. Outliers observed in Kartvelian languages and Indo-European languages with Joshi class 3 may have contributed to this discrepancy in prediction. These observations suggest that increasing the number of data points covering higher spBLEU scores may help mitigate the bias in prediction.

\section{Related Work}

The prediction performance of machine learning algorithms has been mainly explored in two research directions: (1) predict the model performance during the training runtime, and (2) predict the model performance by providing extracted features from the dataset~\cite{xia2020predicting}. 

\paragraph{Performance Prediction During the Training Runtime.}
The former aims to infer and extrapolate the learning curve to approximate training results using evaluation metric measurements \cite{kolachina2012prediction}. \citet{domhan2015speeding} study the quick detection of poor hyper-parameters in probabilistic models after a few steps of Stochastic Gradient Descent (SGD). \citet{adriaensen2024efficient} extrapolate learning curves from a parametric prior using Markov Chain Monte Carlo (MCMC).

\paragraph{Performance Prediction Using Extracted Features.}
The latter aims to predict the model performance by learning a correlation between input features and the final evaluation metric.~\citet{birch2008predicting} identify strong predictive features such as the amount of reordering, the morphological complexity of the target language, and the historical relatedness of the two languages. \citet{xia2020predicting} leverage extracted dataset features and typological database language representations. \citet{ye2021towards} introduce the use of confidence intervals and calibration with various regressor algorithms for reliable performance prediction. \citet{schram2023performance} apply Bayesian matrix factorization for performance prediction on multilingual NLP tasks.
In this work, we focus on exploring the latter.
Existing approaches have shown promise using linear regression and gradient-boosting trees \cite{birch2008predicting,xia2020predicting,srinivasan2021predicting,ye2021towards}.  These studies have considered data size, typological features, and language similarity as factors contributing to the model performance.

\paragraph{Enhancing LLMs through Small Models.}
Recent studies, as compiled in \citet{chen2024role}, have explored leveraging smaller models to complement LLMs across various tasks. A closely related application involves using smaller models to assist with LLM inference and evaluation \cite{kuhn2023semantic,manakul2023selfcheckgpt,wang2024apollo, winata2024metametrics}. In contrast, our work focuses on employing proxy models within a language- and task-agnostic framework to predict LLM performance accurately. This approach offers a cost-effective alternative to fine-tuning and inference during model selection, while achieving high evaluation accuracy and robustness across extreme domain shifts. These include datasets spanning diverse domains, quality levels, and languages, with particular effectiveness demonstrated for extremely LRLs.

\section{Conclusion}
In this paper, we introduce \methodname{}, a novel framework designed to predict the performance of LMs by leveraging proxy models including for LRLs. By utilizing proxy models as substitutes to estimate the performance of the target model, we strategically employ smaller LMs or off-the-shelf models without additional fine-tuning. This framework is highly scalable to multiple proxy models and is task- and language-agnostic, making it applicable to a wide range of downstream NLP tasks. Our approach showcases substantial advancements in prediction accuracy compared to standard baselines and exhibits strong generalization capabilities across varied scenarios.

\section*{Limitations}

This paper focuses on the empirical use of various proxy models without delving into the intricacies of the proxy model selection. Specifically, we do not investigate methods for determining which proxy models are the most effective across different contexts without relying on empirical experimentation. This limitation highlights a significant avenue for future research, which could involve a more systematic approach to identifying the most effective proxy models given estimated LMs and NLP tasks. Nevertheless, the use of proxy models consistently outperforms the absence of such models, demonstrating their importance in performance prediction.

Alternatively, developing robust methodologies for collecting and analyzing relevant past performance records could provide invaluable insights that enhance the generalization and accuracy of our predictive framework. Some performance data may offer greater information gain than others, potentially minimizing the number of performance records required to achieve a more robust and accurate predictor. By establishing a clearer understanding of which performance records yield the most informative insights, we can optimize our approach and improve our overall predictive capabilities. Future research in this area may further explore various data collection strategies and analytical techniques to develop a more comprehensive framework for selecting and utilizing proxy models effectively.

\section*{Ethical Considerations}
We are committed to conducting our evaluations with the utmost standards of transparency and fairness. This commitment involves applying rigorous methodologies that ensure equitable and unbiased assessment processes.

\section*{Acknowledgements}
We extend our sincere gratitude to Viktoria Schram for providing assistance in reproducing Matrix Factorization as our baseline.

\bibliography{custom}

\begin{thebibliography}{57}
\providecommand{\natexlab}[1]{#1}

\bibitem[{Adilazuarda et~al.(2024)Adilazuarda, Cahyawijaya, Aji, Winata, and Purwarianti}]{adilazuarda2024lingualchemy}
Muhammad~Farid Adilazuarda, Samuel Cahyawijaya, Alham~Fikri Aji, Genta~Indra Winata, and Ayu Purwarianti. 2024.
\newblock Lingualchemy: Fusing typological and geographical elements for unseen language generalization.
\newblock \emph{arXiv preprint arXiv:2401.06034}.

\bibitem[{Adriaensen et~al.(2024)Adriaensen, Rakotoarison, M{\"u}ller, and Hutter}]{adriaensen2024efficient}
Steven Adriaensen, Herilalaina Rakotoarison, Samuel M{\"u}ller, and Frank Hutter. 2024.
\newblock Efficient bayesian learning curve extrapolation using prior-data fitted networks.
\newblock \emph{Advances in Neural Information Processing Systems}, 36.

\bibitem[{Allal et~al.(2024)Allal, Lozhkov, Bakouch, von Werra, and Wolf}]{allal2024SmolLM}
Loubna~Ben Allal, Anton Lozhkov, Elie Bakouch, Leandro von Werra, and Thomas Wolf. 2024.
\newblock Smollm - blazingly fast and remarkably powerful.

\bibitem[{Aryabumi et~al.(2024)Aryabumi, Dang, Talupuru, Dash, Cairuz, Lin, Venkitesh, Smith, Marchisio, Ruder et~al.}]{aryabumi2024aya}
Viraat Aryabumi, John Dang, Dwarak Talupuru, Saurabh Dash, David Cairuz, Hangyu Lin, Bharat Venkitesh, Madeline Smith, Kelly Marchisio, Sebastian Ruder, et~al. 2024.
\newblock Aya 23: Open weight releases to further multilingual progress.
\newblock \emph{arXiv preprint arXiv:2405.15032}.

\bibitem[{Birch et~al.(2008)Birch, Osborne, and Koehn}]{birch2008predicting}
Alexandra Birch, Miles Osborne, and Philipp Koehn. 2008.
\newblock Predicting success in machine translation.
\newblock In \emph{Proceedings of the 2008 Conference on Empirical Methods in Natural Language Processing}, pages 745--754.

\bibitem[{Brown et~al.(2020)Brown, Mann, Ryder, Subbiah, Kaplan, Dhariwal, Neelakantan, Shyam, Sastry, Askell et~al.}]{brown2020language}
Tom Brown, Benjamin Mann, Nick Ryder, Melanie Subbiah, Jared~D Kaplan, Prafulla Dhariwal, Arvind Neelakantan, Pranav Shyam, Girish Sastry, Amanda Askell, et~al. 2020.
\newblock Language models are few-shot learners.
\newblock \emph{Advances in neural information processing systems}, 33:1877--1901.

\bibitem[{Cahyawijaya et~al.(2023)Cahyawijaya, Lovenia, Koto, Adhista, Dave, Oktavianti, Akbar, Lee, Shadieq, Cenggoro et~al.}]{cahyawijaya2023nusawrites}
Samuel Cahyawijaya, Holy Lovenia, Fajri Koto, Dea Adhista, Emmanuel Dave, Sarah Oktavianti, Salsabil Akbar, Jhonson Lee, Nuur Shadieq, Tjeng~Wawan Cenggoro, et~al. 2023.
\newblock Nusawrites: Constructing high-quality corpora for underrepresented and extremely low-resource languages.
\newblock In \emph{Proceedings of the 13th International Joint Conference on Natural Language Processing and the 3rd Conference of the Asia-Pacific Chapter of the Association for Computational Linguistics (Volume 1: Long Papers)}, pages 921--945.

\bibitem[{Chen and Varoquaux(2024)}]{chen2024role}
Lihu Chen and Ga{\"e}l Varoquaux. 2024.
\newblock What is the role of small models in the llm era: A survey.
\newblock \emph{arXiv preprint arXiv:2409.06857}.

\bibitem[{Chen and Guestrin(2016)}]{chen2016xgboost}
Tianqi Chen and Carlos Guestrin. 2016.
\newblock Xgboost: A scalable tree boosting system.
\newblock In \emph{Proceedings of the 22nd acm sigkdd international conference on knowledge discovery and data mining}, pages 785--794.

\bibitem[{Christodouloupoulos and Steedman(2015)}]{christodouloupoulos2015massively}
Christos Christodouloupoulos and Mark Steedman. 2015.
\newblock A massively parallel corpus: the bible in 100 languages.
\newblock \emph{Language resources and evaluation}, 49:375--395.

\bibitem[{Cleveland(1979)}]{cleveland1979ze}
William~S Cleveland. 1979.
\newblock Robust locally weighted regression and smoothing scatterplots.
\newblock \emph{J. Am. Stat. Assoc.}, 74(368):829--836.

\bibitem[{Costa-juss{\`a} et~al.(2022)Costa-juss{\`a}, Cross, {\c{C}}elebi, Elbayad, Heafield, Heffernan, Kalbassi, Lam, Licht, Maillard et~al.}]{costa2022no}
Marta~R Costa-juss{\`a}, James Cross, Onur {\c{C}}elebi, Maha Elbayad, Kenneth Heafield, Kevin Heffernan, Elahe Kalbassi, Janice Lam, Daniel Licht, Jean Maillard, et~al. 2022.
\newblock No language left behind: Scaling human-centered machine translation.
\newblock \emph{arXiv preprint arXiv:2207.04672}.

\bibitem[{Domhan et~al.(2015)Domhan, Springenberg, and Hutter}]{domhan2015speeding}
Tobias Domhan, Jost~Tobias Springenberg, and Frank Hutter. 2015.
\newblock Speeding up automatic hyperparameter optimization of deep neural networks by extrapolation of learning curves.
\newblock In \emph{Twenty-fourth international joint conference on artificial intelligence}.

\bibitem[{Dubey et~al.(2024)Dubey, Jauhri, Pandey, Kadian, Al-Dahle, Letman, Mathur, Schelten, Yang, Fan et~al.}]{dubey2024llama}
Abhimanyu Dubey, Abhinav Jauhri, Abhinav Pandey, Abhishek Kadian, Ahmad Al-Dahle, Aiesha Letman, Akhil Mathur, Alan Schelten, Amy Yang, Angela Fan, et~al. 2024.
\newblock The llama 3 herd of models.
\newblock \emph{arXiv preprint arXiv:2407.21783}.

\bibitem[{Dušek et~al.(2014)Dušek, Hajic, Hlaváčová, Novák, Pecina, Rosa, Tamchyna, Urešová, and Zeman}]{duvsek2014machine}
Ondrej Dušek, Jan Hajic, Jaroslava Hlaváčová, Michal Novák, Pavel Pecina, Rudolf Rosa, Ales Tamchyna, Zdenka Urešová, and Daniel Zeman. 2014.
\newblock Machine translation of medical texts in the khresmoi project.
\newblock In \emph{Proceedings of the Ninth Workshop on Statistical Machine Translation}, pages 221--228.

\bibitem[{Fan et~al.(2021)Fan, Bhosale, Schwenk, Ma, El-Kishky, Goyal, Baines, Celebi, Wenzek, Chaudhary et~al.}]{fan2021beyond}
Angela Fan, Shruti Bhosale, Holger Schwenk, Zhiyi Ma, Ahmed El-Kishky, Siddharth Goyal, Mandeep Baines, Onur Celebi, Guillaume Wenzek, Vishrav Chaudhary, et~al. 2021.
\newblock Beyond english-centric multilingual machine translation.
\newblock \emph{Journal of Machine Learning Research}, 22(107):1--48.

\bibitem[{FitzGerald et~al.(2022)FitzGerald, Hench, Peris, Mackie, Rottmann, Sanchez, Nash, Urbach, Kakarala, Singh et~al.}]{fitzgerald2022massive}
Jack FitzGerald, Christopher Hench, Charith Peris, Scott Mackie, Kay Rottmann, Ana Sanchez, Aaron Nash, Liam Urbach, Vishesh Kakarala, Richa Singh, et~al. 2022.
\newblock Massive: A 1m-example multilingual natural language understanding dataset with 51 typologically-diverse languages.
\newblock \emph{arXiv preprint arXiv:2204.08582}.

\bibitem[{Freitag et~al.(2023)Freitag, Mathur, Lo, Avramidis, Rei, Thompson, Kocmi, Blain, Deutsch, Stewart et~al.}]{freitag2023results}
Markus Freitag, Nitika Mathur, Chi-kiu Lo, Eleftherios Avramidis, Ricardo Rei, Brian Thompson, Tom Kocmi, Fr{\'e}d{\'e}ric Blain, Daniel Deutsch, Craig Stewart, et~al. 2023.
\newblock Results of wmt23 metrics shared task: Metrics might be guilty but references are not innocent.
\newblock In \emph{Proceedings of the Eighth Conference on Machine Translation}, pages 578--628.

\bibitem[{Gowda et~al.(2021)Gowda, Zhang, Mattmann, and May}]{gowda2021many}
Thamme Gowda, Zhao Zhang, Chris Mattmann, and Jonathan May. 2021.
\newblock Many-to-english machine translation tools, data, and pretrained models.
\newblock In \emph{Proceedings of the 59th Annual Meeting of the Association for Computational Linguistics and the 11th International Joint Conference on Natural Language Processing: System Demonstrations}, pages 306--316.

\bibitem[{Goyal et~al.(2022)Goyal, Gao, Chaudhary, Chen, Wenzek, Ju, Krishnan, Ranzato, Guzm{\'a}n, and Fan}]{goyal2022flores}
Naman Goyal, Cynthia Gao, Vishrav Chaudhary, Peng-Jen Chen, Guillaume Wenzek, Da~Ju, Sanjana Krishnan, Marc{'}Aurelio Ranzato, Francisco Guzm{\'a}n, and Angela Fan. 2022.
\newblock \href {https://doi.org/10.1162/tacl_a_00474} {The {F}lores-101 evaluation benchmark for low-resource and multilingual machine translation}.
\newblock \emph{Transactions of the Association for Computational Linguistics}, 10:522--538.

\bibitem[{Gu et~al.(2018)Gu, Hassan, Devlin, and Li}]{gu2018universal}
J~Gu, H~Hassan, J~Devlin, and VOK Li. 2018.
\newblock Universal neural machine translation for extremely low resource languages.
\newblock In \emph{Proceedings of the 2018 Conference of the North American Chapter of the Association for Computational Linguistics: Human Language Technologies, Volume 1 (Long Papers)}. Association for Computational Linguistic.

\bibitem[{Haddow and Kirefu(2020)}]{haddow2020pmindia}
Barry Haddow and Faheem Kirefu. 2020.
\newblock Pmindia--a collection of parallel corpora of languages of india.
\newblock \emph{arXiv preprint arXiv:2001.09907}.

\bibitem[{Hu et~al.(2021)Hu, Shen, Wallis, Allen-Zhu, Li, Wang, Wang, and Chen}]{hu2021lora}
Edward~J Hu, Yelong Shen, Phillip Wallis, Zeyuan Allen-Zhu, Yuanzhi Li, Shean Wang, Lu~Wang, and Weizhu Chen. 2021.
\newblock Lora: Low-rank adaptation of large language models.
\newblock \emph{arXiv preprint arXiv:2106.09685}.

\bibitem[{Joshi et~al.(2020)Joshi, Santy, Budhiraja, Bali, and Choudhury}]{joshi2021state}
Pratik Joshi, Sebastin Santy, Amar Budhiraja, Kalika Bali, and Monojit Choudhury. 2020.
\newblock The state and fate of linguistic diversity and inclusion in the nlp world.
\newblock \emph{arXiv preprint arXiv:2004.09095}.

\bibitem[{Kaplan et~al.(2020)Kaplan, McCandlish, Henighan, Brown, Chess, Child, Gray, Radford, Wu, and Amodei}]{kaplan2020scaling}
Jared Kaplan, Sam McCandlish, Tom Henighan, Tom~B Brown, Benjamin Chess, Rewon Child, Scott Gray, Alec Radford, Jeffrey Wu, and Dario Amodei. 2020.
\newblock Scaling laws for neural language models.
\newblock \emph{arXiv preprint arXiv:2001.08361}.

\bibitem[{Khiu et~al.(2024)Khiu, Toossi, Anugraha, Liu, Li, Flores, Roman, Do{\u{g}}ru{\"o}z, and Lee}]{khiu2024predicting}
Eric Khiu, Hasti Toossi, David Anugraha, Jinyu Liu, Jiaxu Li, Juan Armando~Parra Flores, Leandro~Acros Roman, A~Seza Do{\u{g}}ru{\"o}z, and En-Shiun~Annie Lee. 2024.
\newblock Predicting machine translation performance on low-resource languages: The role of domain similarity.
\newblock \emph{arXiv preprint arXiv:2402.02633}.

\bibitem[{Kocmi et~al.(2022)Kocmi, Bawden, Bojar, Dvorkovich, Federmann, Fishel, Gowda, Graham, Grundkiewicz, Haddow, Knowles, Koehn, Monz, Morishita, Nagata, Nakazawa, Nov{\'a}k, Popel, and Popovic}]{Kocmi2022FindingsOT}
Tom Kocmi, Rachel Bawden, Ondrej Bojar, Anton Dvorkovich, Christian Federmann, Mark Fishel, Thamme Gowda, Yvette Graham, Roman Grundkiewicz, Barry Haddow, Rebecca Knowles, Philipp Koehn, Christof Monz, Makoto Morishita, Masaaki Nagata, Toshiaki Nakazawa, Michal Nov{\'a}k, Martin Popel, and Maja Popovic. 2022.
\newblock \href {https://api.semanticscholar.org/CorpusID:256461033} {Findings of the 2022 conference on machine translation (wmt22)}.
\newblock In \emph{Conference on Machine Translation}.

\bibitem[{Kolachina et~al.(2012)Kolachina, Cancedda, Dymetman, and Venkatapathy}]{kolachina2012prediction}
Prasanth Kolachina, Nicola Cancedda, Marc Dymetman, and Sriram Venkatapathy. 2012.
\newblock Prediction of learning curves in machine translation.
\newblock In \emph{Proceedings of the 50th Annual Meeting of the Association for Computational Linguistics (Volume 1: Long Papers)}, pages 22--30.

\bibitem[{Kuhn et~al.(2023)Kuhn, Gal, and Farquhar}]{kuhn2023semantic}
Lorenz Kuhn, Yarin Gal, and Sebastian Farquhar. 2023.
\newblock Semantic uncertainty: Linguistic invariances for uncertainty estimation in natural language generation.
\newblock \emph{arXiv preprint arXiv:2302.09664}.

\bibitem[{Littell et~al.(2017)Littell, Mortensen, Lin, Kairis, Turner, and Levin}]{littell2017uriel}
Patrick Littell, David~R Mortensen, Ke~Lin, Katherine Kairis, Carlisle Turner, and Lori Levin. 2017.
\newblock Uriel and lang2vec: Representing languages as typological, geographical, and phylogenetic vectors.
\newblock In \emph{Proceedings of the 15th Conference of the European Chapter of the Association for Computational Linguistics: Volume 2, Short Papers}, pages 8--14.

\bibitem[{Liu et~al.(2020)Liu, Gu, Goyal, Li, Edunov, Ghazvininejad, Lewis, and Zettlemoyer}]{liu2020multilingual}
Yinhan Liu, Jiatao Gu, Naman Goyal, Xian Li, Sergey Edunov, Marjan Ghazvininejad, Mike Lewis, and Luke Zettlemoyer. 2020.
\newblock Multilingual denoising pre-training for neural machine translation.
\newblock \emph{Transactions of the Association for Computational Linguistics}, 8:726--742.

\bibitem[{Liu et~al.(2021)Liu, Winata, and Fung}]{liu2021continual}
Zihan Liu, Genta~Indra Winata, and Pascale Fung. 2021.
\newblock Continual mixed-language pre-training for extremely low-resource neural machine translation.
\newblock In \emph{Findings of the Association for Computational Linguistics: ACL-IJCNLP 2021}, pages 2706--2718.

\bibitem[{Manakul et~al.(2023)Manakul, Liusie, and Gales}]{manakul2023selfcheckgpt}
Potsawee Manakul, Adian Liusie, and Mark~JF Gales. 2023.
\newblock Selfcheckgpt: Zero-resource black-box hallucination detection for generative large language models.
\newblock \emph{arXiv preprint arXiv:2303.08896}.

\bibitem[{Mohammadshahi et~al.(2022)Mohammadshahi, Nikoulina, B{\'e}rard, Brun, Henderson, and Besacier}]{mohammadshahi2022small}
Alireza Mohammadshahi, Vassilina Nikoulina, Alexandre B{\'e}rard, Caroline Brun, James Henderson, and Laurent Besacier. 2022.
\newblock Small-100: Introducing shallow multilingual machine translation model for low-resource languages.
\newblock In \emph{Proceedings of the 2022 Conference on Empirical Methods in Natural Language Processing}, pages 8348--8359.

\bibitem[{Muennighoff et~al.(2022)Muennighoff, Wang, Sutawika, Roberts, Biderman, Scao, Bari, Shen, Yong, Schoelkopf et~al.}]{muennighoff2022crosslingual}
Niklas Muennighoff, Thomas Wang, Lintang Sutawika, Adam Roberts, Stella Biderman, Teven~Le Scao, M~Saiful Bari, Sheng Shen, Zheng-Xin Yong, Hailey Schoelkopf, et~al. 2022.
\newblock Crosslingual generalization through multitask finetuning.
\newblock \emph{arXiv preprint arXiv:2211.01786}.

\bibitem[{Qi et~al.(2018)Qi, Sachan, Felix, Padmanabhan, and Neubig}]{qi2018and}
Ye~Qi, Devendra Sachan, Matthieu Felix, Sarguna Padmanabhan, and Graham Neubig. 2018.
\newblock When and why are pre-trained word embeddings useful for neural machine translation?
\newblock In \emph{Proceedings of the 2018 Conference of the North American Chapter of the Association for Computational Linguistics: Human Language Technologies, Volume 2 (Short Papers)}, pages 529--535.

\bibitem[{Raffel et~al.(2020)Raffel, Shazeer, Roberts, Lee, Narang, Matena, Zhou, Li, and Liu}]{raffel2020exploring}
Colin Raffel, Noam Shazeer, Adam Roberts, Katherine Lee, Sharan Narang, Michael Matena, Yanqi Zhou, Wei Li, and Peter~J Liu. 2020.
\newblock Exploring the limits of transfer learning with a unified text-to-text transformer.
\newblock \emph{Journal of machine learning research}, 21(140):1--67.

\bibitem[{Rei et~al.(2022)Rei, De~Souza, Alves, Zerva, Farinha, Glushkova, Lavie, Coheur, and Martins}]{rei2022comet}
Ricardo Rei, Jos{\'e}~GC De~Souza, Duarte Alves, Chrysoula Zerva, Ana~C Farinha, Taisiya Glushkova, Alon Lavie, Luisa Coheur, and Andr{\'e}~FT Martins. 2022.
\newblock Comet-22: Unbabel-ist 2022 submission for the metrics shared task.
\newblock In \emph{Proceedings of the Seventh Conference on Machine Translation (WMT)}, pages 578--585.

\bibitem[{Reimers and Gurevych(2019)}]{reimers2019sentencebert}
Nils Reimers and Iryna Gurevych. 2019.
\newblock \href {https://arxiv.org/abs/1908.10084} {Sentence-bert: Sentence embeddings using siamese bert-networks}.
\newblock In \emph{Proceedings of the 2019 Conference on Empirical Methods in Natural Language Processing}. Association for Computational Linguistics.

\bibitem[{Schram et~al.(2023)Schram, Beck, and Cohn}]{schram2023performance}
Viktoria Schram, Daniel Beck, and Trevor Cohn. 2023.
\newblock Performance prediction via bayesian matrix factorisation for multilingual natural language processing tasks.
\newblock In \emph{Proceedings of the 17th Conference of the European Chapter of the Association for Computational Linguistics}, pages 1790--1801.

\bibitem[{Schwenk et~al.(2021)Schwenk, Chaudhary, Sun, Gong, and Guzm{\'a}n}]{schwenk2021wikimatrix}
Holger Schwenk, Vishrav Chaudhary, Shuo Sun, Hongyu Gong, and Francisco Guzm{\'a}n. 2021.
\newblock Wikimatrix: Mining 135m parallel sentences in 1620 language pairs from wikipedia.
\newblock In \emph{Proceedings of the 16th Conference of the European Chapter of the Association for Computational Linguistics: Main Volume}, pages 1351--1361.

\bibitem[{Srinivasan et~al.(2021)Srinivasan, Sitaram, Ganu, Dandapat, Bali, and Choudhury}]{srinivasan2021predicting}
Anirudh Srinivasan, Sunayana Sitaram, Tanuja Ganu, Sandipan Dandapat, Kalika Bali, and Monojit Choudhury. 2021.
\newblock Predicting the performance of multilingual nlp models.
\newblock \emph{arXiv preprint arXiv:2110.08875}.

\bibitem[{Tiedemann(2012)}]{tiedemann2012parallel}
J{\"o}rg Tiedemann. 2012.
\newblock Parallel data, tools and interfaces in opus.
\newblock In \emph{Proceedings of the Eighth International Conference on Language Resources and Evaluation (LREC'12)}, pages 2214--2218.

\bibitem[{Touvron et~al.(2023{\natexlab{a}})Touvron, Lavril, Izacard, Martinet, Lachaux, Lacroix, Rozi{\`e}re, Goyal, Hambro, Azhar et~al.}]{touvron2023llama1}
Hugo Touvron, Thibaut Lavril, Gautier Izacard, Xavier Martinet, Marie-Anne Lachaux, Timoth{\'e}e Lacroix, Baptiste Rozi{\`e}re, Naman Goyal, Eric Hambro, Faisal Azhar, et~al. 2023{\natexlab{a}}.
\newblock Llama: Open and efficient foundation language models.
\newblock \emph{arXiv preprint arXiv:2302.13971}.

\bibitem[{Touvron et~al.(2023{\natexlab{b}})Touvron, Martin, Stone, Albert, Almahairi, Babaei, Bashlykov, Batra, Bhargava, Bhosale et~al.}]{touvron2023llama}
Hugo Touvron, Louis Martin, Kevin Stone, Peter Albert, Amjad Almahairi, Yasmine Babaei, Nikolay Bashlykov, Soumya Batra, Prajjwal Bhargava, Shruti Bhosale, et~al. 2023{\natexlab{b}}.
\newblock Llama 2: Open foundation and fine-tuned chat models.
\newblock \emph{arXiv preprint arXiv:2307.09288}.

\bibitem[{Vaswani et~al.(2017)Vaswani, Shazeer, Parmar, Uszkoreit, Jones, Gomez, Kaiser, and Polosukhin}]{vaswani2017attention}
Ashish Vaswani, Noam Shazeer, Niki Parmar, Jakob Uszkoreit, Llion Jones, Aidan~N Gomez, {\L}ukasz Kaiser, and Illia Polosukhin. 2017.
\newblock Attention is all you need.
\newblock \emph{Advances in neural information processing systems}, 30.

\bibitem[{Wang et~al.(2024)Wang, Chen, Chen, Hu, Wang, Wu, Gao, Wan, Li, and Wang}]{wang2024apollo}
Xidong Wang, Nuo Chen, Junyin Chen, Yan Hu, Yidong Wang, Xiangbo Wu, Anningzhe Gao, Xiang Wan, Haizhou Li, and Benyou Wang. 2024.
\newblock Apollo: Lightweight multilingual medical llms towards democratizing medical ai to 6b people.
\newblock \emph{arXiv preprint arXiv:2403.03640}.

\bibitem[{Winata et~al.(2022)Winata, Wu, Kulkarni, Solorio, and Preo{\c{t}}iuc-Pietro}]{winata2022cross}
Genta Winata, Shijie Wu, Mayank Kulkarni, Thamar Solorio, and Daniel Preo{\c{t}}iuc-Pietro. 2022.
\newblock Cross-lingual few-shot learning on unseen languages.
\newblock In \emph{Proceedings of the 2nd Conference of the Asia-Pacific Chapter of the Association for Computational Linguistics and the 12th International Joint Conference on Natural Language Processing (Volume 1: Long Papers)}, pages 777--791.

\bibitem[{Winata et~al.(2024{\natexlab{a}})Winata, Anugraha, Susanto, Kuwanto, and Wijaya}]{winata2024metametrics}
Genta~Indra Winata, David Anugraha, Lucky Susanto, Garry Kuwanto, and Derry~Tanti Wijaya. 2024{\natexlab{a}}.
\newblock Metametrics: Calibrating metrics for generation tasks using human preferences.
\newblock \emph{arXiv preprint arXiv:2410.02381}.

\bibitem[{Winata et~al.(2024{\natexlab{b}})Winata, Zhang, and Adelani}]{winata2024miners}
Genta~Indra Winata, Ruochen Zhang, and David~Ifeoluwa Adelani. 2024{\natexlab{b}}.
\newblock Miners: Multilingual language models as semantic retrievers.
\newblock \emph{arXiv preprint arXiv:2406.07424}.

\bibitem[{Workshop et~al.(2022)Workshop, Scao, Fan, Akiki, Pavlick, Ili{\'c}, Hesslow, Castagn{\'e}, Luccioni, Yvon et~al.}]{le2023bloom}
BigScience Workshop, Teven~Le Scao, Angela Fan, Christopher Akiki, Ellie Pavlick, Suzana Ili{\'c}, Daniel Hesslow, Roman Castagn{\'e}, Alexandra~Sasha Luccioni, Fran{\c{c}}ois Yvon, et~al. 2022.
\newblock Bloom: A 176b-parameter open-access multilingual language model.
\newblock \emph{arXiv preprint arXiv:2211.05100}.

\bibitem[{Xia et~al.(2020)Xia, Anastasopoulos, Xu, Yang, and Neubig}]{xia2020predicting}
Mengzhou Xia, Antonios Anastasopoulos, Ruochen Xu, Yiming Yang, and Graham Neubig. 2020.
\newblock Predicting performance for natural language processing tasks.
\newblock In \emph{Proceedings of the 58th Annual Meeting of the Association for Computational Linguistics}, pages 8625--8646.

\bibitem[{Ye et~al.(2021)Ye, Liu, Fu, and Neubig}]{ye2021towards}
Zihuiwen Ye, Pengfei Liu, Jinlan Fu, and Graham Neubig. 2021.
\newblock Towards more fine-grained and reliable nlp performance prediction.
\newblock In \emph{Proceedings of the 16th Conference of the European Chapter of the Association for Computational Linguistics: Main Volume}, pages 3703--3714.

\bibitem[{Yong et~al.(2024)Yong, Menghini, and Bach}]{yong2024lexc}
Zheng-Xin Yong, Cristina Menghini, and Stephen~H Bach. 2024.
\newblock Lexc-gen: Generating data for extremely low-resource languages with large language models and bilingual lexicons.
\newblock \emph{arXiv preprint arXiv:2402.14086}.

\bibitem[{Zhang et~al.(2020)Zhang, Williams, Titov, and Sennrich}]{zhang2020improving}
Biao Zhang, Philip Williams, Ivan Titov, and Rico Sennrich. 2020.
\newblock Improving massively multilingual neural machine translation and zero-shot translation.
\newblock In \emph{Proceedings of the 58th Annual Meeting of the Association for Computational Linguistics}, pages 1628--1639.

\bibitem[{Zhu et~al.(2023)Zhu, Liu, Dong, Xu, Huang, Kong, Chen, and Li}]{zhu2023multilingual}
Wenhao Zhu, Hongyi Liu, Qingxiu Dong, Jingjing Xu, Shujian Huang, Lingpeng Kong, Jiajun Chen, and Lei Li. 2023.
\newblock Multilingual machine translation with large language models: Empirical results and analysis.
\newblock \emph{arXiv preprint arXiv:2304.04675}.

\bibitem[{Zoph et~al.(2016)Zoph, Yuret, May, and Knight}]{zoph2016transfer}
Barret Zoph, Deniz Yuret, Jonathan May, and Kevin Knight. 2016.
\newblock Transfer learning for low-resource neural machine translation.
\newblock In \emph{Proceedings of the 2016 Conference on Empirical Methods in Natural Language Processing}, pages 1568--1575.

\end{thebibliography}

\appendix

\section{Dataset Features Calculation}
\label{sec:app-detail-dataset-feat}

In this section, we will describe in detail how the dataset features were computed.
\begin{itemize}
    \item \textit{Train size}: The number of dataset entries used for training specified NLP task.
    \item \textit{Vocab size}: The number of unique tokens in the dataset after tokenization and preprocessing, where we use SentencePiece.
    \item \textit{Average Sentence Length}: The average number of tokens per sentence in the dataset.
    \item \textit{Word Overlap}:
    $$\dfrac{|T_1 \cap T_2|}{|T_1| + |T_2|}$$
    where $T_1$ and $T_2$ correspond to the unique tokens of two datasets.
    \item \textit{Type-Token Ratio (TTR)}: The ratio of unique tokens to the total number of tokens in the dataset. This is given by:
    $$
    \text{TTR} = \dfrac{|V|}{\sum_{i=1}^{N} |s_i|}
    $$
    where $|V|$ is the number of unique tokens, and $\sum_{i=1}^{N} |s_i|$ is the total number of tokens in all $N$ sentences.
    
    \item \textit{TTR Distance}: The distance between the TTRs of two datasets, calculated as:
    $$
    D_{\text{TTR}} = \left( 1 - \dfrac{\text{TTR}_1}{\text{TTR}_2} \right)^2
    $$
    where $\text{TTR}_1$ and $\text{TTR}_2$ are the TTR values for two datasets.
    
    \item \textit{Jensen-Shannon Divergence (JSD)}: Measures the divergence between two token distributions. For two token distributions $P$ and $Q$, the JSD is calculated as:
    $$
    \text{JSD}(P, Q) = \dfrac{1}{2} \left[ \text{KL}(P \parallel M) + \text{KL}(Q \parallel M) \right]
    $$
    where $M = \dfrac{1}{2} (P + Q)$ and $\text{KL}$ is the Kullback-Leibler divergence:
    $$
    \text{KL}(P \parallel Q) = \sum_{x} P(x) \log \dfrac{P(x)}{Q(x)}
    $$
    
    \item \textit{TF-IDF Cosine Similarity}: The cosine similarity between two datasets based on their TF-IDF representations. The cosine similarity is given by:
    $$
    \text{Cosine Similarity}(A, B) = \dfrac{A \cdot B}{\|A\| \|B\|}
    $$
    where $A$ and $B$ are the TF-IDF vectors of the two datasets.
    
    \item \textit{Sentence-BERT Similarity}: The cosine similarity between the mean sentence embeddings of two datasets obtained from Sentence-BERT \cite{reimers2019sentencebert}. It is computed as:
    $$
    \text{Cosine Similarity}(E_A, E_B) = \dfrac{E_A \cdot E_B}{\|E_A\| \|E_B\|}
    $$
    where $E_A$ and $E_B$ are the mean sentence embeddings for datasets $A$ and $B$, respectively.
\end{itemize}

\section{Experimental Details}
\label{sec:appendix}

\begin{table*}[!t]
\centering
\resizebox{0.96\textwidth}{!}{
    \begin{tabular}{llc}
    \toprule   
    \textbf{Datasets} & \multicolumn{1}{l}{\textbf{Languages Under Study}} & \textbf{Domain} \\
    \midrule
    \multicolumn{3}{l}{\textbf{English-centric Dataset}} \\
    $\quad$FLoRes-200~\cite{costa2022no} & afr, amh, arz, bak, bel, ceb, cym, deu, dik, ewe, & Multi-domain \\ & fao, fra, guj, hau, hne, hye, ibo, ind, jav, kan, \\ & kat, kaz, khm, kin, kir, kor, lmo, ltz, mar, mri, \\ &  mya, oci, pan, plt, rus, sin, sna, snd, som, ssw,   \\ & tam, tat, tgl, tuk, wol, xho, vie, yor, zho, zul & \\ 
    $\quad$MT560~\cite{gowda2021many} \\
    $\quad\quad$Europarl & deu & Wiki \\ 
    $\quad\quad$DGT & fra & Government \\ 
    $\quad\quad$Joshua Indian Corpus & tam & Wiki \\ 
    $\quad\quad$Neulab TED Talk~\cite{qi2018and} & bel, deu, fra, hye, kat, kaz, kor, mar, mya, rus, & TED \\
    & tam, vie, zho & \\
    $\quad\quad$News Commentary & deu, fra, rus, zho & News \\
    $\quad\quad$News-test WMT22 ~\cite{Kocmi2022FindingsOT} & deu, rus, zho & News \\ 
    $\quad\quad$General-test WMT22 ~\cite{Kocmi2022FindingsOT} & deu, fra, kaz, rus, zho & News \\ 
    $\quad\quad$OPUS100~\cite{zhang2020improving} & afr, amh, bel, cym, deu, fra, guj, hau, hye, ibo,  & Multi-domain\\
    & ind, kan, kat, kaz, khm, kin, kir, kor, mar, mya,  & \\
    & oci, pan, rus, sin, tam, tat, tuk, vie, xho, yor, \\ & zho, zul \\ 
    $\quad\quad$OPUS Bible & afr, amh, ceb, deu, dik, ewe, fra, guj, hye, ind, & Religion \\ 
    $\quad\quad$\cite{christodouloupoulos2015massively} &  kan, kor, mar, mri, mya, plt, rus, sin, sna, som, \\ & ssw, tam, tgl, vie, wol, xho, zho, zul \\ 
    $\quad\quad$OPUS OpenSubtitles & afr, hye, kat, kor, sin, tgl, vie, zho & Movies \\ 
    $\quad\quad$OPUS Tatoeba~\cite{tiedemann2012parallel} & afr, amh, arz, bak, bel, ceb, fao, hau, hye, ibo, & Conversational \\ & ind, jav, kan, kat, kaz, khm, kin, kir, kor, lmo, \\ & ltz, mar, mri, mya, pan, sna, tat, tgl, tuk, vie, \\ & yor, zul \\ 
    $\quad\quad$OPUS Tanzil~\cite{tiedemann2012parallel}& amh, deu, fra, hau, ind, kor, rus, snd, som, tat, & Religion\\ & zho \\
    $\quad\quad$OPUS Gnome~\cite{tiedemann2012parallel} & fao, mri, som, zho & Technology \\ 
    $\quad\quad$OPUS GlobalVoices~\cite{tiedemann2012parallel} & amh, kor, mya & News \\ 
    $\quad\quad$OPUS Wikipedia Health~\cite{tiedemann2012parallel} & fra, kor, rus, vie & Health \\ 
    $\quad\quad$OPUS Antibiotic~\cite{tiedemann2012parallel} & fra & Health \\ 
    $\quad\quad$OPUS Tico~\cite{tiedemann2012parallel} & fra, rus & Health \\ 
    $\quad\quad$OPUS Vaccination~\cite{tiedemann2012parallel} & fra & Health \\ 
    $\quad\quad$OPUS Ubuntu~\cite{tiedemann2012parallel} & vie, zho & Technology \\ 
    $\quad\quad$OPUS EU Bookshop~\cite{tiedemann2012parallel} & cym & Government \\ 
    $\quad\quad$OPUS SPC & afr & Government \\ 
    $\quad\quad$OPUS Memat & xho & Medicine \\ 
    $\quad\quad$OPUS XhosaNavy & xho & Government \\ 
    $\quad\quad$OPUS KDE4 & hne & Technology \\ 
    $\quad\quad$OPUS infopankki & som & Immigration \\ 
    $\quad\quad$OPUS TLDR & zho & General \\
    $\quad\quad$UN & rus & Government \\
    $\quad\quad$lindat khresmoi~\cite{duvsek2014machine} & fra & Health \\ 
    $\quad\quad$PMIndia~\cite{haddow2020pmindia} & guj, kan, mar, pan, sin, tam & Government \\ 
    $\quad\quad$Wiki Titles & guj, kaz, tam & Wiki \\ $\quad\quad$WikiMatrix~\cite{schwenk2021wikimatrix} & arz, bak, bel, ceb, fao, fra, ind, jav, kat, kaz, & Wiki \\
    & kor, lmo, ltz, mar, oci, sin, tam, tat, tgl, vie, & \\ & zho \\ \midrule
    \multicolumn{3}{l}{\textbf{Many-to-Many Languages Dataset}} \\
    $\quad$NusaTranslation~\cite{cahyawijaya2023nusawrites} & 
    bew, btk, ind, jav, mad, mak, min, sun & Social Media 
    \\ \bottomrule
    \end{tabular}
}
\caption{List of datasets under study covering 50 different languages. We only opt for 50 out of 500 languages available in the MT560 dataset, 50 out of 200 languages available in the FLoRes-200 dataset, and 8 out of 12 languages available in the NusaTranslation dataset.}
\label{tab:datasets}
\end{table*}

\subsection{Languages Under Study}
\label{sec:more-info-langs}
We list all the languages used in the training from the MT560~\cite{gowda2021many} and NusaTranslation~\cite{cahyawijaya2023nusawrites} datasets in Table \ref{tab:lang-mt560} and Table \ref{tab:lang-nusawrites}, respectively. The language code follows $^{*}$ISO639-3 coding. All languages are also complemented by their $^{\dagger}$rarity taxonomy based on \cite{joshi2021state} into two vitality classes: 0-2→low resource language (LRL), 3-4→mid resource language (MRL), and 5→high resource language (HRL). We also provide information about whether the language was part of the pre-trained M2M100 model dataset to highlight the model knowledge coverage.

\subsection{Models}
\label{paragraph-models-details}
The details on the proxy models we use in MT experiments are as follows:
\begin{itemize}
    \item Transformer (100M)~\cite{vaswani2017attention}: a standard encoder-decoder transformer-based model with 6 encoder layers and 6 decoder layers with an embedding dimension of 512. We train the model from randomly initialized parameters with the training set.
    \item SMaLL-100 (330M)~\cite{mohammadshahi2022small}:\footnote{SMaLL-100 (330M) is taken from~\url{https://github.com/alirezamshi/small100}.} a distilled version of the M2M100 (12B) model. We utilize the model in two ways: fine-tuned on training data and zero-shot inference.
    \item M2M100 (No FT)~\cite{fan2021beyond}:\footnote{M2M100 (1.2B) is taken from~\url{https://github.com/facebookresearch/fairseq/tree/main/examples/m2m_100}.} a pre-trained estimated model of M2M100 (1.2B) without any fine-tuning. We run the model in a zero-shot fashion.
    \item NLLB (No FT)~\cite{costa2022no}:\footnote{NLLB (1.3B) is taken from~\url{https://github.com/facebookresearch/fairseq/tree/nllb}.} a pre-trained estimated model of NLLB-200 Distilled (1.3B) without any fine-tuning. We run the model in a zero-shot fashion.
\end{itemize}

The details on the proxy models we use in intent classification and slot filling experiments are as follows:

\begin{itemize}
    \item SmolLM (135M and 360M)~\cite{allal2024SmolLM}:\footnote{SmolLM 135M and 360M are taken from \url{https://huggingface.co/HuggingFaceTB/SmolLM-135M} and \url{https://huggingface.co/HuggingFaceTB/SmolLM-360M} respectively} a series of small LMs built on Cosmo-Corpus, a meticulously curated high-quality training dataset. SmolLM models have shown promising results when compared to other models in their size categories across various benchmarks testing common sense reasoning and world knowledge.
    \item BLOOMZ (560M)~\cite{muennighoff2022crosslingual}: a decoder model fine-tuned from BLOOM \cite{le2023bloom} on a cross-lingual task mixture (xP3) and finds the resulting models capable of cross-lingual generalization to unseen tasks and languages.
\end{itemize}

\begin{table*}[!ht]
\centering
\small
\resizebox{\textwidth}{!}{
    \begin{tabular}{@{\extracolsep{2pt}}ccccccc}
    \toprule   
     \textbf{Language} & \textbf{Language Code}$^{*}$ & \textbf{Family} & \textbf{Joshi Class}$^{\dagger}$ & \textbf{Vitality}$^{\dagger}$ & \textbf{Seen by M2M100} & \textbf{Covered by COMET-22} \\ 
    \midrule
    Afrikaans & afr & indo-european & 3 & MRL & \cmark & \cmark \\
    Amharic & amh & afro-asiatic & 2 & LRL & \cmark & \cmark \\
    Armenian & hye & indo-european & 1 & LRL & \cmark & \cmark \\
    Bashkir & bak & turkic & 1 & LRL & \cmark & \xmark \\
    Belarusian & bel & indo-european & 3 & MRL & \cmark & \cmark \\
    Burmese & mya & sino-tibetan & 1 & LRL & \cmark & \cmark \\
    Cebuano & ceb & austronesian & 3 & MRL & \cmark &\xmark \\
    Chhattisgarhi & hne & indo-european & 0 & LRL & \xmark & \xmark \\
    Chinese & zho & sino-tibetan & 5 & HRL & \cmark & \cmark \\
    Dinka & dik & nilo-saharan & 1 & LRL & \xmark & \xmark \\
    Egyptian Arabic & arz & afro-asiatic & 3 & MRL & \cmark & \xmark \\
    Ewe & ewe & niger-congo & 1 & LRL & \xmark & \xmark \\
    Faroese & fao & indo-european & 1 & LRL & \xmark & \xmark \\
    Georgian & kat & kartvelian & 3 & MRL & \cmark & \cmark \\
    German & deu & indo-european & 5 & HRL & \cmark & \cmark \\
    Gujarati & guj & indo-european & 1 & LRL & \cmark & \cmark \\
    French & fra & indo-european & 5 & HRL & \cmark & \cmark \\
    Hausa & hau & afro-asiatic & 2 & LRL & \cmark & \cmark \\
    Igbo & ibo & niger-congo & 1 & LRL & \cmark & \xmark \\
    Indonesian & ind & austronesian & 3 & MRL & \cmark & \cmark \\
    Javanese & jav & austronesian & 1 & LRL & \cmark & \cmark \\
    Kannada & kan & dravidian & 1 & LRL & \cmark & \cmark \\
    Kazakh & kaz & turkic & 3 & MRL & \cmark & \cmark \\
    Khmer & khm & austro-asiatic & 1 & LRL & \cmark & \cmark \\
    Kirghiz & kir & turkic & 1 & LRL & \xmark & \cmark \\
    Kinyarwanda & kin & niger-congo & 1 & LRL & \xmark & \xmark \\
    Korean & kor & koreanic & 4 & MRL & \cmark & \cmark \\
    Lombard & lmo & indo-european & 1 & LRL & \xmark & \xmark\\
    Luxembourgish & ltz & indo-european & 1 & LRL & \cmark & \xmark \\
    Malagasy & plt & austronesian & 1 & LRL & \cmark & \cmark \\
    Maori & mri & austronesian & 1 & LRL & \xmark & \xmark \\
    Marathi & mar & indo-european & 2 & LRL & \cmark & \cmark \\
    Occitan & oci & indo-european & 1 & LRL & \cmark & \xmark\\
    Punjabi & pan & indo-european & 2 & LRL & \cmark & \cmark \\
    Russian & rus & indo-european & 4 & MRL & \cmark & \cmark \\
    Shona & sna & niger-congo & 1 & LRL & \xmark & \xmark \\
    Sindhi & snd & indo-european & 1 & LRL & \cmark & \cmark \\
    Sinhala & sin & indo-european & 0 & LRL & \cmark & \cmark \\
    Somali & som & afro-asiatic & 1 & LRL & \cmark & \cmark \\
    Swati & ssw & niger-congo & 1 & LRL & \cmark & \xmark \\
    Tagalog & tgl & austronesian & 3 & MRL & \cmark & \xmark \\
    Tamil & tam & dravidian & 3 & MRL & \cmark& \cmark  \\
    Tatar & tat & turkic & 1 & LRL & \xmark & \xmark \\
    Turkmen & tuk & turkic & 1 & LRL & \xmark & \xmark \\
    Vietnamese & vie & austro-asiatic & 4 & MRL & \cmark & \cmark \\
    Welsh & cym & indo-european & 1 & LRL & \cmark & \cmark \\
    Wolof & wol & niger-congo & 2 & LRL & \cmark & \xmark \\
    Xhosa & xho & niger-congo & 2 & LRL & \cmark & \cmark \\
    Yoruba & yor & niger-congo & 2 & LRL & \cmark & \xmark \\
    Zulu & zul & niger-congo & 2 & LRL & \cmark & \xmark\\
    \bottomrule
    \end{tabular}
}
\caption{List of languages from the English-centric dataset on MT task, including their rarity category mapping, an indication of whether they are involved in the pre-training process for M2M100, and another indication of whether they are covered by COMET-22. Note that all languages are involved in the pre-training process for NLLB.
} 
\label{tab:lang-mt560}
\end{table*}

\begin{table*}[!ht]
\centering
\small
\resizebox{0.8\textwidth}{!}{
    \begin{tabular}{@{\extracolsep{2pt}}ccccc}
    \toprule   
     \textbf{Language} & \textbf{Language Code}$^{*}$ & \textbf{Family} & \textbf{Joshi Class}$^{\dagger}$ & \textbf{Vitality}$^{\dagger}$ \\ 
    \midrule
    Afrikaans & afr & indo-european & 3 & MRL \\
    Amharic & amh & afro-asiatic & 2 & LRL \\
    Arabic & ara & afro-asiatic & 5 & HRL \\
    Azerbaijani            & aze  & turkic & 1 & LRL               \\
    Bengali                & ben & indo-european & 3 & MRL                 \\
    Welsh                  & cym & indo-european & 1 & LRL                 \\ 
    Danish                 & dan & indo-european & 3 & MRL                 \\ 
    German                 & deu & indo-european & 5 & HRL                 \\ 
    Greek                  & ell & indo-european & 3 & MRL                 \\ 
    English                & eng & indo-european & 5 & HRL                 \\ 
    Spanish                & spa & indo-european & 5 & HRL                 \\ 
    Persian                & fas & indo-european & 4 & MRL                 \\ 
    Finnish                & fin & uralic & 4 & MRL                 \\ 
    French                 & fra & indo-european & 5 & HRL                 \\ 
    Hebrew                 & heb & afro-asiatic & 3 & MRL                 \\ 
    Hindi                  & hin & indo-european & 4 & MRL                 \\ 
    Hungarian              & hun & uralic & 4 & MRL                 \\ 
    Armenian               & hye & indo-european & 1 & LRL                 \\ 
    Indonesian             & ind & austronesian & 3 & MRL                 \\ 
    Icelandic              & isl & indo-european & 2 & LRL                 \\ 
    Italian                & ita & indo-european & 4 & MRL                 \\ 
    Japanese               & jpn & japonic & 5 & HRL                 \\ 
    Javanese               & jav & austronesian & 1 & LRL                 \\ 
    Georgian               & kat & kartvelian & 3 & MRL                 \\ 
    Khmer                  & khm & austro-asiatic & 1 & LRL                 \\ 
    Kannada                & kan & dravidian & 1 & LRL                 \\ 
    Korean                 & kor & koreanic & 4 & MRL                 \\ 
    Latvian                & lav & indo-european & 3 & MRL                 \\ 
    Malayalam              & mal & dravidian & 1 & LRL                 \\ 
    Mongolian              & mon & mongolic & 1 & LRL                 \\ 
    Malay                  & msa & austronesian & 3 & MRL                 \\ 
    Burmese                & mya & sino-tibetan & 1 & LRL                 \\ 
    Norwegian Bokmål       & nob & indo-european & 1 & LRL                 \\ 
    Dutch                  & nld & indo-european & 4 & MRL                 \\ 
    Polish                 & pol & indo-european & 4 & MRL                 \\ 
    Portuguese             & por & indo-european & 4 & MRL                 \\ 
    Romanian               & ron & indo-european & 3 & MRL                 \\ 
    Russian                & rus & indo-european & 4 & MRL                 \\ 
    Slovenian              & slv & indo-european & 3 & MRL                 \\ 
    Albanian               & sqi & indo-european & 1 & LRL                 \\ 
    Swedish                & swe & indo-european & 4 & MRL                 \\ 
    Swahili                & swa & niger-congo & 2 & LRL                 \\ 
    Tamil                  & tam & dravidian & 3 & MRL                 \\ 
    Telugu                 & tel & dravidian & 1 & LRL                 \\ 
    Thai                   & tha & kra-dai & 3 & MRL                 \\ 
    Tagalog                & tgl & austronesian & 3 & MRL                 \\ 
    Turkish                & tur & turkic & 4 & MRL                 \\ 
    Urdu                   & urd & indo-european & 3 & MRL                 \\ 
    Vietnamese             & vie & indo-european & 4 & MRL                 \\ 
    Chinese (Simplified)   & zho & sino-tibetan & 5 & HRL                 \\ 
    Chinese (Traditional)  & zho & sino-tibetan & 5 & HRL                 \\ 
    
    \bottomrule
    \end{tabular}
}
\caption{List of languages from the MASSIVE dataset for intent classification and slot filling, including their rarity category mapping.
} 
\label{tab:lang-massive}
\end{table*}

\subsection{Hyper-parameters}
\label{hyperparam}

\paragraph{LM}

Each fine-tuning and evaluation for LMs is done with an NVIDIA Tesla V100 32GB GPU.

The hyper-parameters used during fine-tuning from the English-centric and Many-to-Many Languages datasets for MT task are listed in Table \ref{hyperparam-small100}, \ref{hyperparam-m2m100}, \ref{hyperparam-nllb}, and \ref{hyperparam-transformer} for SMaLL100, M2M100, NLLB, and Transformer models, respectively.

The common hyper-parameters used during fine-tuning for intent and slot classification tasks are listed in Table \ref{tab:hyperparam-intent-slot} for SmolLM-135M, SmolLM-360M, BLOOMZ-560M, Aya23-8B, and LLaMA3-8B respectively. The learning rate used for SmolLM-135M, SmolLM-360M, and BLOOMZ-560M is $1e^{-5}$, while the learning rate used for Aya23-8B and LLaMA3-8B is $1e^{-4}$. Since we are fine-tuning using SFT, we use the ``default" template for SmolLM-135M, SmolLM-360M, and BLOOMZ-560M, while we use ``cohere" for Aya23-8B and ``llama3" for LLaMA3-8B. Additionally, we use LoRA for both Aya23-8B and LLaMA3-8B.

\paragraph{Regressor}

Each regressor is trained on an AMD Ryzen Threadripper 2990WX with 128 GB of RAM and 16 threads. Regressors' hyper-parameters used are provided in Table \ref{hyperparam-xgb-mt-eng},  \ref{hyperparam-xgb-mt-many}, \ref{hyperparam-xgb-intent-slot}, \ref{hyperparam-poly}, \ref{hyperparam-lgbm}, and \ref{hyperparam-mf} for XGBoost, Poly2/Poly3, LGBM, and MF, respectively. These hyper-parameters were obtained based on the best cross-validation RMSE score on training using 10 folds.

\begin{table*}[!ht]
\centering
\resizebox{\textwidth}{!}{
    \begin{tabular}{@{\extracolsep{2pt}}cccccccc}
    \toprule   
     \textbf{Language} & \textbf{Language Code}$^{*}$ & \textbf{Family} & \textbf{Joshi Class}$^{\dagger}$ & \textbf{Vitality}$^{\dagger}$ & \textbf{Seen by M2M100} & \textbf{Seen by NLLB}\\ 
    \midrule
    Indonesian & ind & austronesian & 3 & MRL & \cmark & \cmark \\
    Javanese & jav & austronesian & 1 & LRL & \cmark & \cmark \\
    Betawi & bew & creole & 0 & LRL & \xmark & \xmark \\
    Batak & btk & austronesian & 0 & LRL & \xmark & \xmark \\
    Madurese & mad & austronesian & 0 & LRL & \xmark & \xmark \\
    Makassarese & mak & austronesian & 0 & LRL & \xmark & \xmark \\
    Minangkabau & min & austronesian & 0 & LRL & \xmark & \cmark \\
    Sundanese & sun & austronesian & 1 & LRL & \cmark & \cmark \\
    \bottomrule
    \end{tabular}
}
\caption{List of languages from the Many-to-Many Languages dataset on MT task along with their rarity category mapping and an indication of whether they are included in the pre-training process for each respective model.
} 
\label{tab:lang-nusawrites}
\end{table*}

\begin{table}[!th]
\centering
\resizebox{.49\textwidth}{!}{
    \begin{tabular}{l|cc}
        \toprule
        \textbf{Hyper-parameter} & \textbf{English-centric} & \textbf{Many-to-Many Langs.} \\
        \midrule
        Encoder Layers & 12 & 12 \\
        Decoder Layers & 3 & 3 \\
        Encoder Embed Dim & 1024 & 1024 \\
        Decoder Embed Dim & 1024 & 1024 \\
        Encoder FFN Embed Dim & 4096 & 4096 \\
        Decoder FFN Embed Dim & 4096 & 4096 \\
        Encoder Attention Heads & 16 & 16 \\
        Decoder Attention Heads & 16 & 16 \\
        Encoder Layerdrop & 0.05 & 0.05 \\
        Decoder Layerdrop & 0.05 & 0.05 \\
        Optimizer & Adam & Adam \\
        Adam Eps & 1e-6 & 1e-6 \\
        Adam Betas & (0.9, 0.98) & (0.9, 0.98) \\
        Patience & 6 & 6 \\
        Batch Size & 16 & 16 \\
        Dropout & 0.1 & 0.1 \\
        Attention Dropout & 0.1 & 0.1 \\
        ReLU Dropout & 0.1 & 0.1 \\
        Weight Decay & 0.0 & 0.0 \\
        Label Smoothing & 0.1 & 0.1 \\
        Clip Norm & 1.0 & 1.0 \\
        Learning Rate & 0.0001 & 0.0003 \\
        Max Tokens (per GPU) & 1,000 & 1,000 \\
        \bottomrule
    \end{tabular}
}
\caption{List of hyper-parameters used for SMaLL100 with English-centric and Many-to-Many Languages datasets.}
\label{hyperparam-small100}
\end{table}

\begin{table}[!th]
\centering
\resizebox{.49\textwidth}{!}{
    \begin{tabular}{l|cc}
        \toprule
        \textbf{Hyper-parameter} & \textbf{English-centric} & \textbf{Many-to-Many Langs.} \\
        \midrule
        Encoder Layers & 24 & 24 \\
        Decoder Layers & 24 & 24 \\
        Encoder Embed Dim & 1024 & 1024 \\
        Decoder Embed Dim & 1024 & 1024 \\
        Encoder FFN Embed Dim & 8192 & 8192 \\
        Decoder FFN Embed Dim & 8192 & 8192 \\
        Encoder Attention Heads & 16 & 16 \\
        Decoder Attention Heads & 16 & 16 \\
        Encoder Layerdrop & 0.05 & 0.05 \\
        Decoder Layerdrop & 0.05 & 0.05 \\
        Optimizer & Adam & Adam \\
        Adam Eps & 1e-6 & 1e-6 \\
        Adam Betas & (0.9, 0.98) & (0.9, 0.98) \\
        Patience & 6 & 6 \\
        Batch Size & 32 & 32 \\
        Dropout & 0.1 & 0.1 \\
        Attention Dropout & 0.1 & 0.1 \\
        ReLU Dropout & 0.1 & 0.1 \\
        Weight Decay & 0.0 & 0.0 \\
        Label Smoothing & 0.1 & 0.1 \\
        Clip Norm & 0.0 & 0.0 \\
        Learning Rate & 0.0002 & 0.0002 \\
        Max Tokens (per GPU) & 1,792 & 1,792 \\
        \bottomrule
    \end{tabular}
}
\caption{List of hyper-parameters used for M2M100 with English-centric and Many-to-Many Languages datasets.}
\label{hyperparam-m2m100}
\end{table}

\begin{table}[!th]
\centering
\resizebox{.49\textwidth}{!}{
    \begin{tabular}{l|c}
        \toprule
        \textbf{Hyper-parameter} & \textbf{Value} \\
        \midrule
        Cutoff Length & 256 \\
        Preprocessing Num Workers & 16 \\
        Train Batch Size & 1 \\
        Gradient Accumulation Steps & 2 \\
        Epochs & 3 \\
        Learning Rate Scheduler & cosine \\
        Warmup Ratio & 0.1 \\
        Eval Batch Size & 1\\
        Eval Steps & 500 \\
        \bottomrule
    \end{tabular}
}
\caption{List of common hyper-parameters used for fine-tuning SmolLM-135M, SmolLM-360M, BLOOMZ-560M, Aya23-8B, and LLaMA3-8B on intent and slot classification task.}
\label{tab:hyperparam-intent-slot}
\end{table}

\subsection{Regressor Dataset Sizes}

We provide the details of the regressor's training and test set size in Table \ref{tab:regressor-train-test-size}.

\section{More Detailed Results}
\label{sec:ensemble-breakdown}

\begin{table*}[!th]
\centering
\resizebox{0.8\textwidth}{!}{
    \begin{tabular}{lrrrrr}
    \toprule   
    \textbf{Models} & \multicolumn{2}{c}{\textbf{Random}} & \multicolumn{2}{c}{\textbf{LOLO}} \\
    & \multicolumn{1}{c}{\textbf{M2M100}$\downarrow$} & \multicolumn{1}{c}{\textbf{NLLB}$\downarrow$} & \multicolumn{1}{c}{\textbf{M2M100}$\downarrow$} & \multicolumn{1}{c}{\textbf{NLLB}$\downarrow$} & \textbf{Avg.} \\
    \midrule 
    \multicolumn{5}{l}{NLPerf~\cite{xia2020predicting} with different regressors} & \\ \midrule
    XGBoost & 7.69 \scriptsize $\pm$ 0.59 & 7.73 \scriptsize $\pm$ 0.08 & 9.20 \scriptsize $\pm$ 0.40 & 12.92 \scriptsize $\pm$ 0.54 & 9.16 \\
    Poly2~\cite{khiu2024predicting} & 11.21 \scriptsize $\pm$ 0.49 & 16.23 \scriptsize $\pm$ 3.51 & 15.55 \scriptsize $\pm$ 0.00 & 43.02 \scriptsize $\pm$ 0.00 & 34.62 \\
    Poly3~\cite{khiu2024predicting} & 11.00 \scriptsize $\pm$ 0.49 & 15.64 \scriptsize $\pm$ 2.59 & 62.29 \scriptsize $\pm$ 0.00 & 236.29 \scriptsize $\pm$ 0.00 & 66.97 \\
    LGBM~\cite{ye2021towards} & 7.88 \scriptsize $\pm$ 0.65 & 8.15 \scriptsize $\pm$ 0.19 & 9.71 \scriptsize $\pm$ 0.17 & 12.81 \scriptsize $\pm$ 0.23 & 9.62 \\
    \midrule
    \multicolumn{5}{l}{\methodname{} (Ours)$^{\ddagger}$ with different proxy models} & \\ \midrule
    Transformer & 4.68 \scriptsize $\pm$ 0.41 & 7.22 \scriptsize $\pm$ 0.25 & 6.18 \scriptsize $\pm$ 0.30 & 11.78 \scriptsize $\pm$ 0.50 & 7.31 \\
    SMaLL-100 & \underline{4.07} \scriptsize $\pm$ 0.31 & 6.33 \scriptsize $\pm$ 0.11 & \underline{4.59} \scriptsize $\pm$ 0.24 & 10.33 \scriptsize $\pm$ 0.41 & 6.04 \\
    SMaLL-100 (No FT) & 5.27 \scriptsize $\pm$ 0.50 & 6.04 \scriptsize $\pm$ 0.32 & 6.28 \scriptsize $\pm$ 0.32 & 10.94 \scriptsize $\pm$ 0.50 & 6.95 \\
    Estimated Model (No FT) & 5.23 \scriptsize $\pm$ 0.54 & \underline{4.15} \scriptsize $\pm$ 0.24 & 6.18 \scriptsize $\pm$ 0.31 & \underline{5.42} \scriptsize $\pm$ 0.27 & \underline{5.38} \\
    Ensemble$^\dagger$ & \textbf{3.21} \scriptsize $\pm$ 0.29 & \textbf{3.68} \scriptsize $\pm$ 0.36 & \textbf{3.74} \scriptsize $\pm$ 0.20 & \textbf{4.94} \scriptsize $\pm$ 0.29 & \textbf{4.01}
    \\ \bottomrule
    \end{tabular}
}
\caption{English-centric test results using spBLEU in average RMSE $\pm$ standard deviation (\textbf{lower is better)}. \textbf{Bold} numbers indicate the best performance, while \underline{underlined} numbers represent the second-best performance. The columns show the setting and estimated model. ``No FT" denotes ``no fine-tuning" and the model inference is done in a zero-shot fashion. Avg represents the average of the results across the row. $^{\ddagger}$The reported results use XGBoost as the regressor. $^\dagger$Ensemble denotes combining all four proxy models.}
\label{tab:app-results-english-centric-detail}
\end{table*}

\begin{table*}[!th]
\centering
\resizebox{0.99\textwidth}{!}{
    \begin{tabular}{lrrrrr}
    \toprule   
    \textbf{Models} & \multicolumn{2}{c}{\textbf{Random}} & \multicolumn{2}{c}{\textbf{LOLO}} \\
    & \multicolumn{1}{c}{\textbf{M2M100}$\downarrow$} & \multicolumn{1}{c}{\textbf{NLLB}$\downarrow$} & \multicolumn{1}{c}{\textbf{M2M100}$\downarrow$} & \multicolumn{1}{c}{\textbf{NLLB}$\downarrow$} & \textbf{Avg.} \\
    \midrule 
    \multicolumn{5}{l}{NLPerf~\cite{xia2020predicting} with different regressors} & \\ \midrule
    XGBoost & 0.0717 \scriptsize $\pm$ 0.0018
    & 0.0556 \scriptsize $\pm$ 0.0038
    & 0.1130 \scriptsize $\pm$ 0.0054
    & 0.1007 \scriptsize $\pm$ 0.0046  
    & 0.0825 \\
    Poly2~\cite{khiu2024predicting} & 0.1325 \scriptsize $\pm$ 0.0014
    & 0.1503 \scriptsize $\pm$ 0.0019
    & 0.1258 \scriptsize $\pm$ 0.0000
    & 0.1176 \scriptsize $\pm$ 0.0000
    & 0.1316 \\
    Poly3~\cite{khiu2024predicting} & 0.1325 \scriptsize $\pm$ 0.0014
    & 0.1520 \scriptsize $\pm$ 0.0019
    & 0.1258 \scriptsize $\pm$ 0.0000
    & 0.1176 \scriptsize $\pm$ 0.0000
    & 0.1320 \\
    LGBM~\cite{ye2021towards} & 0.0752 \scriptsize $\pm$ 0.0025
    & 0.0600 \scriptsize $\pm$ 0.0033
    & 0.1147 \scriptsize $\pm$ 0.0000
    & 0.1019 \scriptsize $\pm$ 0.0000
    & 0.0880 \\
    \midrule
    \multicolumn{5}{l}{\methodname{} (Ours)$^{\ddagger}$ with different proxy models} & \\ \midrule
    Transformer & 0.0543 \scriptsize $\pm$ 0.0009 & 0.0492 \scriptsize $\pm$ 0.0041
    & 0.0837 \scriptsize $\pm$ 0.0042
    & 0.0957 \scriptsize $\pm$ 0.0031
    & 0.0707 \\
    SMaLL-100 & \underline{0.0328} \scriptsize $\pm$ 0.0012
    & 0.0431 \scriptsize $\pm$ 0.0033
    & \underline{0.0547} \scriptsize $\pm$ 0.0027
    & 0.0824 \scriptsize $\pm$ 0.0030
    & \underline{0.0533} \\
    SMaLL-100 (No FT) & 0.0454 \scriptsize $\pm$ 0.0024
    & 0.0393 \scriptsize $\pm$ 0.0033
    & 0.0778 \scriptsize $\pm$ 0.0030
    & 0.0967 \scriptsize $\pm$ 0.0034
    & 0.0648 \\
    Estimated Model (No FT) & 0.0460 \scriptsize $\pm$ 0.0017
    & \underline{0.0348} \scriptsize $\pm$ 0.0016
    & 0.0777 \scriptsize $\pm$ 0.0034
    & \textbf{0.0607} \scriptsize $\pm$ 0.0034
    & 0.0548 \\
    Ensemble$^\dagger$ & \textbf{0.0289} \scriptsize $\pm$ 0.0012
    & \textbf{0.0293} \scriptsize $\pm$ 0.0013
    & \textbf{0.0454} \scriptsize $\pm$ 0.0019
    & \underline{0.0613} \scriptsize $\pm$ 0.0031
    & \textbf{0.0412}
    \\ \bottomrule
    \end{tabular}
}
\caption{English-centric MT test results using COMET-22 in average RMSE $\pm$ standard deviation (\textbf{lower is better)}. \textbf{Bold} numbers indicate the best performance, while \underline{underlined} numbers represent the second-best performance. The columns show the setting and estimated model. ``No FT" denotes ``no fine-tuning" and the model inference is done in a zero-shot fashion. There are no results for the Unseen setting since COMET-22 score does not cover the unseen languages to M2M100. Furthermore, all languages are covered by NLLB in English-centric dataset. Avg represents the average of the results across the row. $^{\ddagger}$The reported results use XGBoost as the regressor. $^\dagger$Ensemble denotes combining all four proxy models, the detailed breakdown of this result can be seen in Section~\ref{sec:ensemble-breakdown} in the Appendix.}
\label{tab:results-english-centric-comet}
\end{table*}

\begin{table*}[!th]
\centering
\resizebox{0.8\textwidth}{!}{
    \begin{tabular}{lrrrrr}
    \toprule   
    \textbf{Models} & \multicolumn{2}{c}{\textbf{Random}} & \multicolumn{2}{c}{\textbf{LOLO}} \\
    & \multicolumn{1}{c}{\textbf{M2M100}$\downarrow$} & \multicolumn{1}{c}{\textbf{NLLB}$\downarrow$} & \multicolumn{1}{c}{\textbf{M2M100}$\downarrow$} & \multicolumn{1}{c}{\textbf{NLLB}$\downarrow$} & \textbf{Avg.}\\
    \midrule
    \multicolumn{3}{l}{NLPerf~\cite{xia2020predicting} with different regressors} & &  \\ \midrule
    XGBoost& 2.45 \scriptsize $\pm$ 0.30 & \textbf{1.11} \scriptsize $\pm$ 0.06 & 7.83 \scriptsize $\pm$ 0.23 & 8.28 \scriptsize $\pm$ 0.31 & 4.94 \\
    Poly2 ~\cite{khiu2024predicting} & 4.70 \scriptsize $\pm$ 0.40 & 4.68 \scriptsize $\pm$ 0.51 & 7.07 \scriptsize $\pm$ 0.00 & 7.90 \scriptsize $\pm$ 0.00 & 6.09 \\
    Poly3 ~\cite{khiu2024predicting} & 4.60 \scriptsize $\pm$ 0.41 & 4.64 \scriptsize $\pm$ 0.49 & 7.26 \scriptsize $\pm$ 0.00 & 8.01 \scriptsize $\pm$ 0.00 & 6.13 \\
    LGBM ~\cite{ye2021towards} & 2.66 \scriptsize $\pm$ 0.22 & 1.76 \scriptsize $\pm$ 0.29 & 7.91 \scriptsize $\pm$ 0.01 & 8.06 \scriptsize $\pm$ 0.00 & 5.10 \\
    MF~\cite{schram2023performance} & 3.65 \scriptsize $\pm$ 0.26 & 2.60 \scriptsize $\pm$ 0.39 & 7.08 \scriptsize $\pm$ 0.23 & 7.14 \scriptsize $\pm$ 0.22 & 5.12 \\
    \midrule
    \multicolumn{3}{l}{\methodname{}$^{\ddagger}$ (Ours) with different proxy models} & & \\ \midrule
    Transformer & 2.56 \scriptsize $\pm$ 0.43 & 1.70 \scriptsize $\pm$ 0.20 & 5.65 \scriptsize $\pm$ 0.23 & 6.24 \scriptsize $\pm$ 0.34 & 4.04 \\
    SMaLL-100 & 2.56 \scriptsize $\pm$ 0.33 & 1.65 \scriptsize $\pm$ 0.44 & \underline{4.85} \scriptsize $\pm$ 0.36 & \underline{5.14} \scriptsize $\pm$ 0.46 & \underline{3.55} \\ 
    SMaLL-100 (No FT) & 2.44 \scriptsize $\pm$ 0.21 & 1.34 \scriptsize $\pm$ 0.38 & 6.93 \scriptsize $\pm$ 0.34 & 7.25 \scriptsize $\pm$ 0.37 & 4.49 \\ 
    Estimated Model (No FT) & \textbf{2.38} \scriptsize $\pm$ 0.36 & \underline{1.27} \scriptsize $\pm$ 0.03 & 5.10 \scriptsize $\pm$ 0.28 & 5.50 \scriptsize $\pm$ 0.26 & 3.56 \\
    Ensemble$^\dagger$ & \underline{2.41} \scriptsize $\pm$ 0.28 & 1.56 \scriptsize $\pm$ 0.35 & \textbf{3.73} \scriptsize $\pm$ 0.23 & \textbf{3.79} \scriptsize $\pm$ 0.19 & \textbf{2.90}
    \\ \bottomrule
    \end{tabular}
}
\caption{Many-to-Many Languages test results in average RMSE $\pm$ standard deviation (\textbf{lower is better)}. \textbf{Bold} numbers indicate the best performance, while \underline{underlined} numbers represent the second-best performance. The columns show the setting and estimated model. ``No FT" denotes ``no fine-tuning". Avg represents the average of the results across the row. $^{\ddagger}$The reported results are experiments using XGBoost regressor. $^{\dagger}$Ensemble denotes combining all four proxy models.}
\label{tab:app-results-many-to-many-detail}
\end{table*}

\begin{table*}[!t]
\centering
\resizebox{.65\textwidth}{!}{
    \begin{tabular}{lrrr}
    \toprule   
    \textbf{Models} & \multicolumn{1}{c}{\textbf{Unseen}} & \multicolumn{2}{c}{\textbf{Cross-Dataset}} \\
    & \textbf{M2M100}$\downarrow$ & \textbf{M2M100}$\downarrow$ & \textbf{NLLB}$\downarrow$ \\
    \midrule
    \multicolumn{3}{l}{NLPerf~\cite{xia2020predicting} with different regressors}\\ \midrule
    XGBoost & 8.26 \scriptsize $\pm$ 0.53 & 12.90 \scriptsize $\pm$ 1.01 & 12.02 \scriptsize $\pm$ 1.02 \\
    Poly2 ~\cite{khiu2024predicting} & 9.51 \scriptsize $\pm$ 0.00 & 11.02 \scriptsize $\pm$ 0.00 & 8.97 \scriptsize $\pm$ 0.00  \\
    Poly3 ~\cite{khiu2024predicting} & 9.64 \scriptsize $\pm$ 0.00 & 11.06 \scriptsize $\pm$ 0.00 & 10.98 \scriptsize $\pm$ 0.00  \\
    LGBM ~\cite{ye2021towards} & 9.56 \scriptsize $\pm$ 0.97 & 9.34 \scriptsize $\pm$ 0.00 & 10.58 \scriptsize $\pm$ 0.00  \\
    \midrule
    \multicolumn{3}{l}{\methodname{}$^{\ddagger}$ (Ours) with different proxy models}\\ \midrule
    Transformer & 6.71 \scriptsize $\pm$ 0.37 & 9.31 \scriptsize $\pm$ 0.00 & 9.70 \scriptsize $\pm$ 0.00   \\
    SMaLL-100 & \underline{4.87} \scriptsize $\pm$ 0.34 & \underline{4.52} \scriptsize $\pm$ 0.05 & \underline{6.87} \scriptsize $\pm$ 0.11  \\ 
    SMaLL-100 (No FT) & 6.22 \scriptsize $\pm$ 0.38 & 8.26 \scriptsize $\pm$ 0.59 & 9.50 \scriptsize $\pm$ 0.04  \\ 
    Estimated Model (No FT) & 5.90 \scriptsize $\pm$ 0.40 & 9.95 \scriptsize $\pm$ 0.40 & 7.20 \scriptsize $\pm$ 0.24 \\
    Ensemble$^\dagger$ & \textbf{4.48} \scriptsize $\pm$ 0.23 & \textbf{4.35} \scriptsize $\pm$ 0.02 & \textbf{5.03} \scriptsize $\pm$ 0.48
    \\ \bottomrule
    \end{tabular}
}
\caption{Unseen and Cross-Dataset MT test results on English-centric dataset in average RMSE $\pm$ standard deviation (\textbf{lower is better)}. \textbf{Bold} numbers indicate the best performance, while \underline{underlined} numbers represent the second-best performance. The columns show the setting and estimated model. ``No FT" denotes ``no fine-tuning". We only show M2M100 results for the Unseen setting since NLLB covers all languages in the English-centric dataset. $^{\ddagger}$The reported results for the Unseen setting use XGBoost, while the Cross-Dataset experiments use LGBM. $^{\dagger}$Ensemble denotes combining all four proxy models.}
\label{tab:app-results-robust-detail}
\end{table*}

\begin{table*}[!th]
\centering
\resizebox{0.9\textwidth}{!}{
    \begin{tabular}{lrrrrr}
    \toprule   
    \textbf{Models} & \multicolumn{2}{c}{\textbf{Random}} & \multicolumn{2}{c}{\textbf{LOLO}} \\
    & \multicolumn{1}{c}{\textbf{Aya-23}$\downarrow$} & \multicolumn{1}{c}{\textbf{LLaMA3}$\downarrow$} & \multicolumn{1}{c}{\textbf{Aya-23}$\downarrow$} & \multicolumn{1}{c}{\textbf{LLaMA3}$\downarrow$} & \textbf{Avg.} \\
    \midrule 
    \multicolumn{5}{l}{NLPerf~\cite{xia2020predicting} with different regressors} & \\ \midrule
    XGBoost &
    0.0761 \scriptsize $\pm$ 0.0008 &
    0.0191 \scriptsize $\pm$ 0.0008 &
    0.1573 \scriptsize $\pm$ 0.0039 &
    0.0581 \scriptsize $\pm$ 0.0021 & 0.0777 \\
    Poly2~\cite{khiu2024predicting} & 0.1996 \scriptsize $\pm$ 0.0017 & 0.0979 \scriptsize $\pm$ 0.0033 & 0.2075 \scriptsize $\pm$ 0.0000 & 0.0918 \scriptsize $\pm$ 0.0000 & 0.1492 \\
    Poly3~\cite{khiu2024predicting} & 0.1990 \scriptsize $\pm$ 0.0011 & 0.0969 \scriptsize $\pm$ 0.0015 & 0.2191 \scriptsize $\pm$ 0.0000 & 0.0925 \scriptsize $\pm$ 0.0000 & 0.1519 \\
    LGBM~\cite{ye2021towards} & 0.0839 \scriptsize $\pm$ 0.0030 & 0.0198 \scriptsize $\pm$ 0.0023 & 0.1545 \scriptsize $\pm$ 0.0000 & 0.0558 \scriptsize $\pm$ 0.0006 & 0.0785 \\
    \midrule
    \multicolumn{5}{l}{\methodname{} (Ours)$^{\ddagger}$ with different proxy models} & \\ \midrule
    SmolLM (135M) & 0.0676 \scriptsize $\pm$ 0.0013 & 0.0171 \scriptsize $\pm$ 0.0008 & 0.1273 \scriptsize $\pm$ 0.0033 & 0.0455 \scriptsize $\pm$ 0.0020 & 0.0644 \\
    SmolLM (360M) & \textbf{0.0604} \scriptsize $\pm$ 0.0015 & \textbf{0.0157} \scriptsize $\pm$ 0.0003 & \underline{0.1118} \scriptsize $\pm$ 0.0031 & \textbf{0.0441} \scriptsize $\pm$ 0.0017 & \textbf{0.0580} \\
    BLOOMZ (560M) & 0.0692 \scriptsize $\pm$ 0.0025 & 0.0179 \scriptsize $\pm$ 0.0009 & 0.1283 \scriptsize $\pm$ 0.0038 & 0.0482 \scriptsize $\pm$ 0.0023 & 0.0659 \\
    Ensemble$^\dagger$ & \underline{0.0609} \scriptsize $\pm$ 0.0010 & \underline{0.0164} \scriptsize $\pm$ 0.0008 & \textbf{0.1112} \scriptsize $\pm$ 0.0032 & \underline{0.0442} \scriptsize $\pm$ 0.0019 & \underline{0.0582}
    \\ \bottomrule
    \end{tabular}
}
\caption{Intent classification results using accuracy in average RMSE $\pm$ standard deviation (\textbf{lower is better)}. \textbf{Bold} numbers indicate the best performance, while \underline{underlined} numbers represent the second-best performance. Avg represents the average of the results across the row. $^{\ddagger}$The reported results use XGBoost as the regressor. $^\dagger$Ensemble denotes combining all four proxy models.}
\label{tab:app-results-intent-detail}
\end{table*}

\begin{table*}[!th]
\centering
\resizebox{0.9\textwidth}{!}{
    \begin{tabular}{lrrrrr}
    \toprule   
    \textbf{Models} & \multicolumn{2}{c}{\textbf{Random}} & \multicolumn{2}{c}{\textbf{LOLO}} \\
    & \multicolumn{1}{c}{\textbf{Aya-23}$\downarrow$} & \multicolumn{1}{c}{\textbf{LLaMA3}$\downarrow$} & \multicolumn{1}{c}{\textbf{Aya-23}$\downarrow$} & \multicolumn{1}{c}{\textbf{LLaMA3}$\downarrow$} & \textbf{Avg.} \\
    \midrule 
    \multicolumn{5}{l}{NLPerf~\cite{xia2020predicting} with different regressors} & \\ \midrule
    XGBoost & 0.0693 \scriptsize $\pm$ 0.0009 & 0.0548 \scriptsize $\pm$ 0.0034 & 0.1219 \scriptsize $\pm$ 0.0035 & 0.1093 \scriptsize $\pm$ 0.0027 & 0.0888 \\
    Poly2~\cite{khiu2024predicting} & 0.1396 \scriptsize $\pm$ 0.0008 & 0.1412 \scriptsize $\pm$ 0.0017 & 0.1418 \scriptsize $\pm$ 0.0000 & 0.1414 \scriptsize $\pm$ 0.0000 & 0.1410 \\
    Poly3~\cite{khiu2024predicting} & 0.1393 \scriptsize $\pm$ 0.0008 & 0.1401 \scriptsize $\pm$ 0.0017 & 0.1448 \scriptsize $\pm$ 0.0000 & 0.1413 \scriptsize $\pm$ 0.0000 & 0.1414 \\
    LGBM~\cite{ye2021towards} & 0.0692 \scriptsize $\pm$ 0.0017 & 0.0557 \scriptsize $\pm$ 0.0029 & 0.1218 \scriptsize $\pm$ 0.0000 & 0.1152 \scriptsize $\pm$ 0.0000 & 0.0905 \\
    \midrule
    \multicolumn{5}{l}{\methodname{} (Ours)$^{\ddagger}$ with different proxy models} & \\ \midrule
    SmolLM (135M) & 0.0618 \scriptsize $\pm$ 0.0018 & 0.0538 \scriptsize $\pm$ 0.0022 & 0.1004 \scriptsize $\pm$ 0.0028 & 0.0953 \scriptsize $\pm$ 0.0024 & 0.0778 \\
    SmolLM (360M) & \textbf{0.0562} \scriptsize $\pm$ 0.0011 & \textbf{0.0506} \scriptsize $\pm$ 0.0023 & \underline{0.0844} \scriptsize $\pm$ 0.0024 & \textbf{0.0868} \scriptsize $\pm$ 0.0018 & \textbf{0.0695} \\
    BLOOMZ (560M) & 0.0618 \scriptsize $\pm$ 0.0027 & 0.0540 \scriptsize $\pm$ 0.0019 & 0.1023 \scriptsize $\pm$ 0.0033 & 0.0995 \scriptsize $\pm$ 0.0028 & 0.0794 \\
    Ensemble$^\dagger$ & \underline{0.0561} \scriptsize $\pm$ 0.0016 & \underline{0.0508} \scriptsize $\pm$ 0.0021 & \textbf{0.0830} \scriptsize $\pm$ 0.0022 & \underline{0.0884} \scriptsize $\pm$ 0.0025 & \underline{0.0696}
    \\ \bottomrule
    \end{tabular}
}
\caption{Slot filling results using micro-F1 score in average RMSE $\pm$ standard deviation (\textbf{lower is better)}. \textbf{Bold} numbers indicate the best performance, while \underline{underlined} numbers represent the second-best performance. Avg represents the average of the results across the row. $^{\ddagger}$The reported results use LGBM as the regressor. $^\dagger$Ensemble denotes combining all four proxy models.}
\label{tab:app-results-slot-detail}
\end{table*}

\begin{figure*}[!th]
    \centering
    \includegraphics[width=1\linewidth]{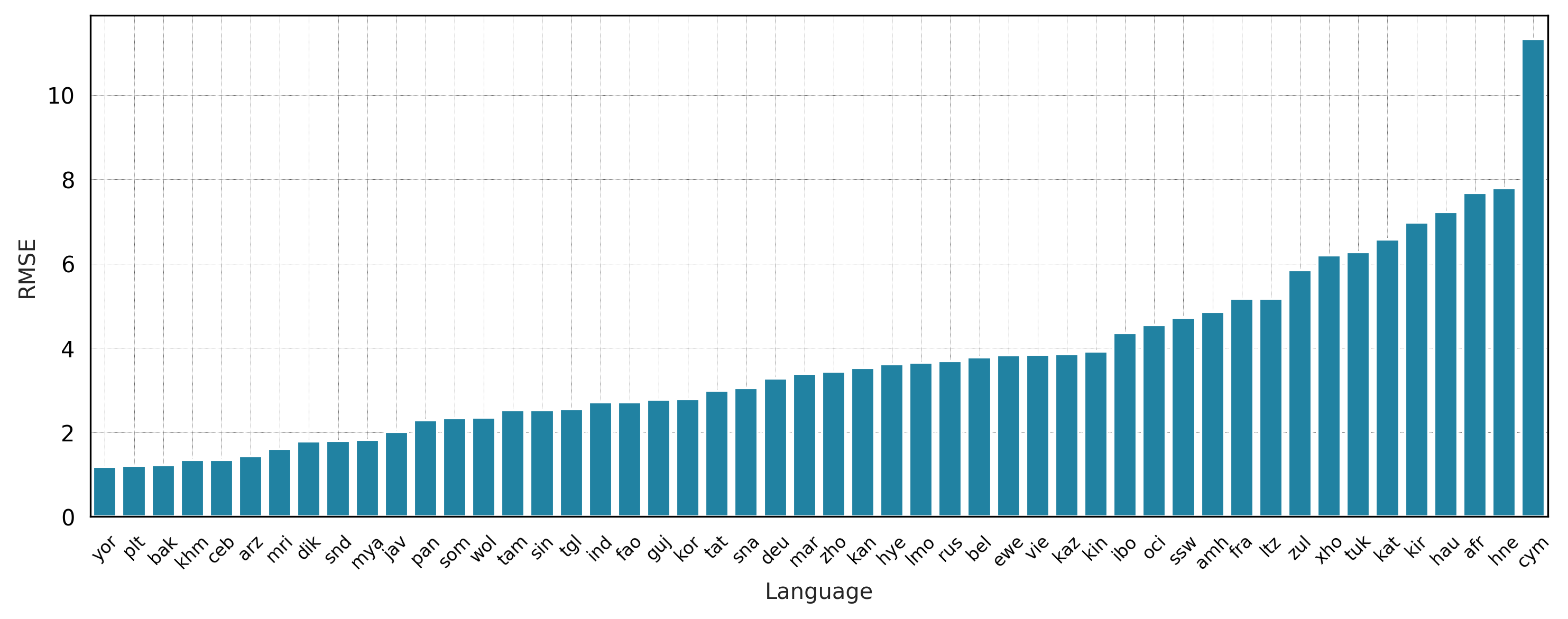}
    \caption{Detailed results of XGBoost with \methodname{} Ensemble on the \textbf{M2M100} model under the \textbf{LOLO} setting using the \textbf{English-centric} dataset on \textbf{MT task} from Table \ref{tab:app-results-english-centric-detail} per languages.}
    \label{fig:lolo-m2m100-mt560-by-lang-appendix}
\end{figure*}

\begin{figure*}[!th]
    \centering
    \includegraphics[width=1\linewidth]{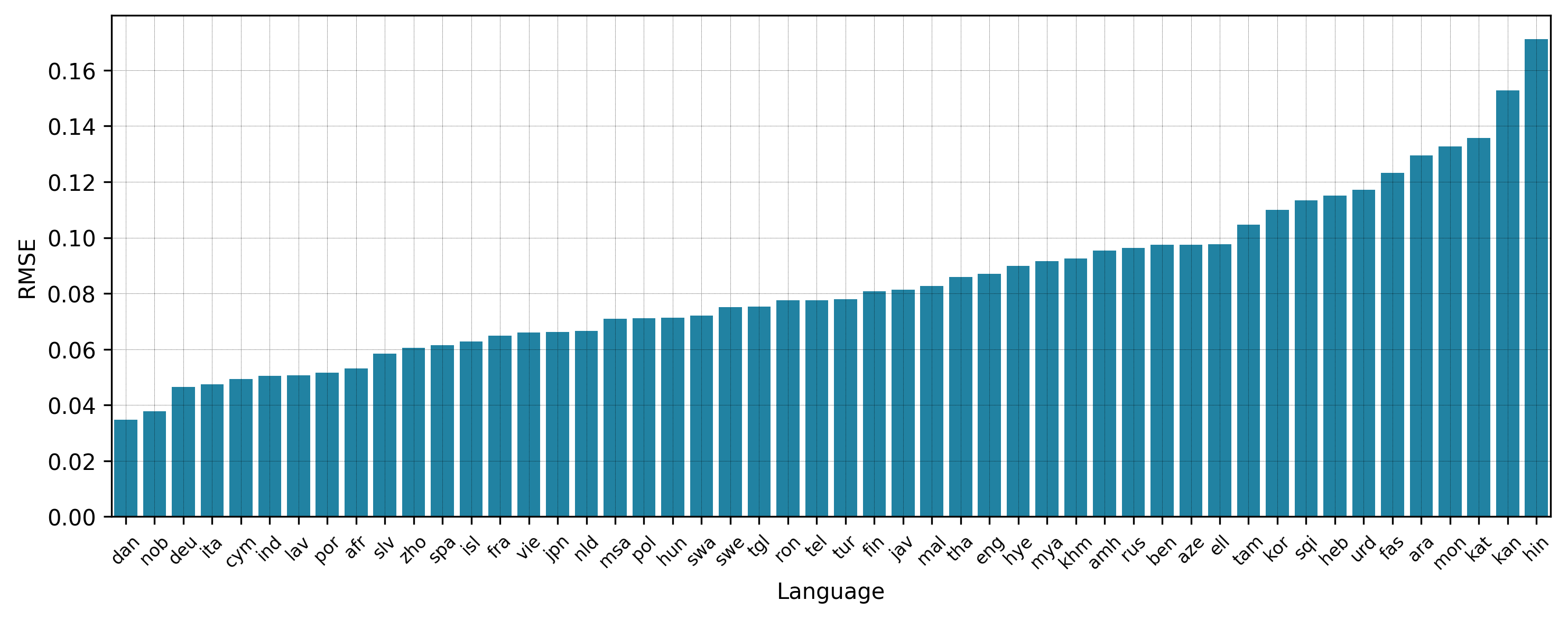}
    \caption{Detailed results of LGBM with \methodname{} Ensemble on the \textbf{Aya} model under the \textbf{LOLO} setting on \textbf{slot classification} from Table \ref{tab:app-results-slot-detail} per languages.}
    \label{fig:lolo-aya-slot-by-lang-appendix}
\end{figure*}

We provide detailed results for Table \ref{tab:results-mt}, \ref{tab:results-intent-slot-combined}, and \ref{fig:result-robust} by providing the standard errors of the predictions. The mapping of vitality, Joshi class, and language family follows the classifications in Table \ref{tab:lang-mt560} and \ref{tab:lang-nusawrites}. The mapping of all languages in Table \ref{fig:lolo-m2m100-mt560-by-lang-appendix} until \ref{fig:lolo-aya-slot-by-lang-appendix}.

\section{Feature Importances}

\begin{figure*}[!th]
    \centering
    \includegraphics[width=0.9\linewidth]{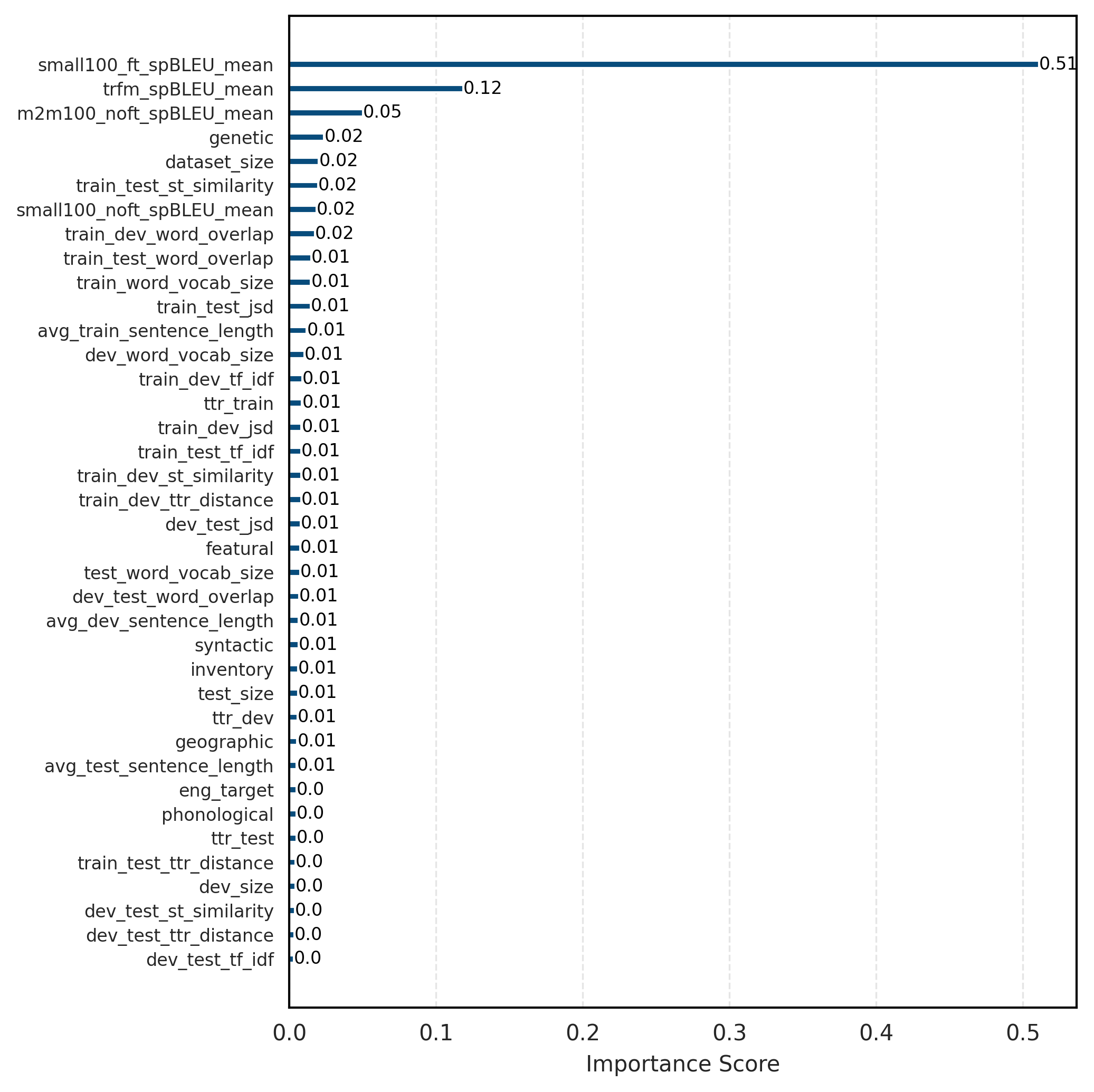}
    \caption{Feature importance analysis of XGBoost with \methodname{} Ensemble on \textbf{MT task} using \textbf{M2M100} model using the \textbf{English-centric} dataset.}
    \label{fig:m2m100-mt560-feat}
\end{figure*}

\begin{figure*}[!th]
    \centering
    \includegraphics[width=0.9\linewidth]{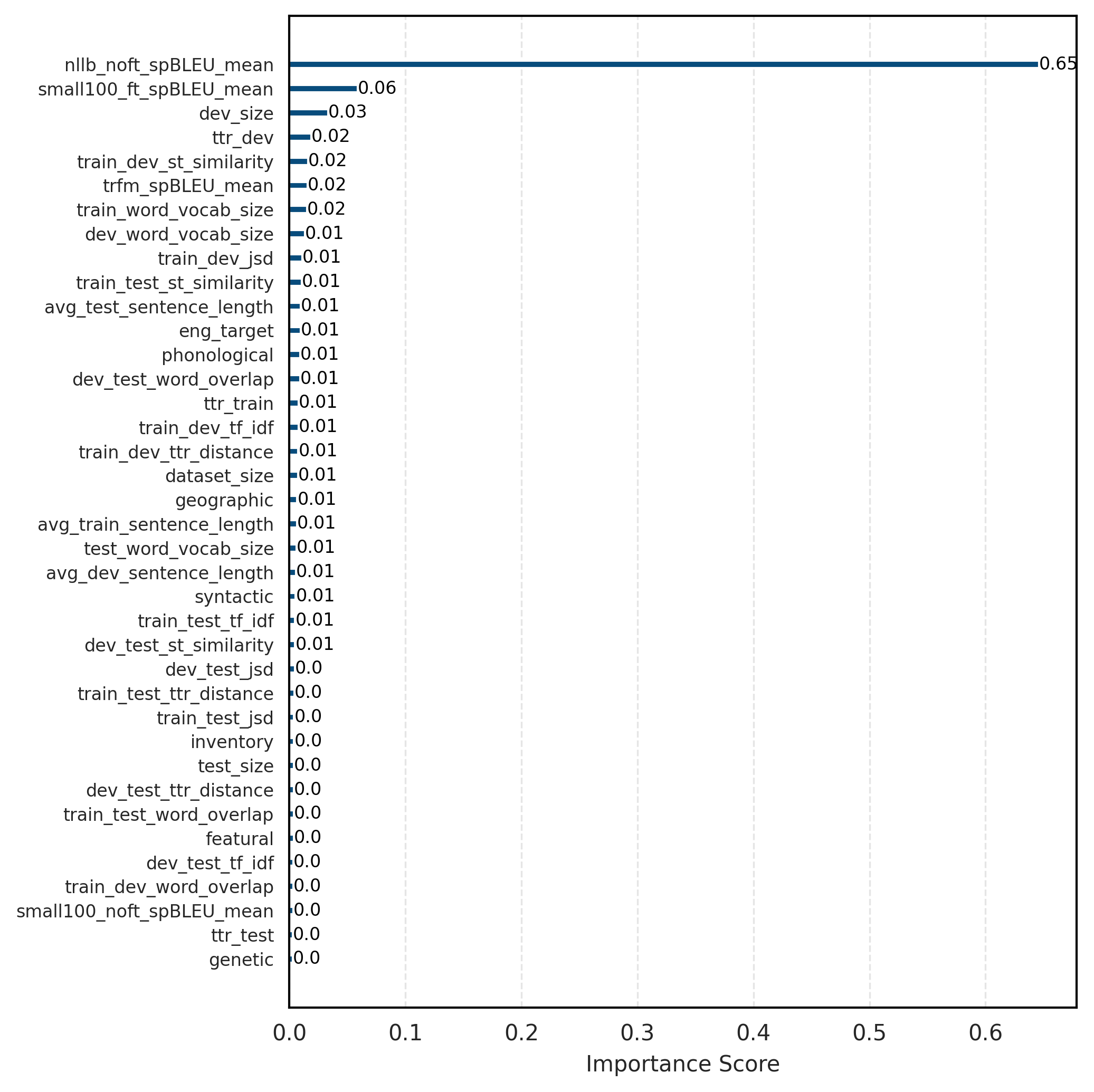}
    \caption{Feature importance analysis of XGBoost with \methodname{} Ensemble on \textbf{MT task} using \textbf{NLLB} model using the \textbf{English-centric} dataset.}
    \label{fig:nllb-mt560-feat}
\end{figure*}

\begin{figure*}[!th]
    \centering
    \includegraphics[width=0.9\linewidth]{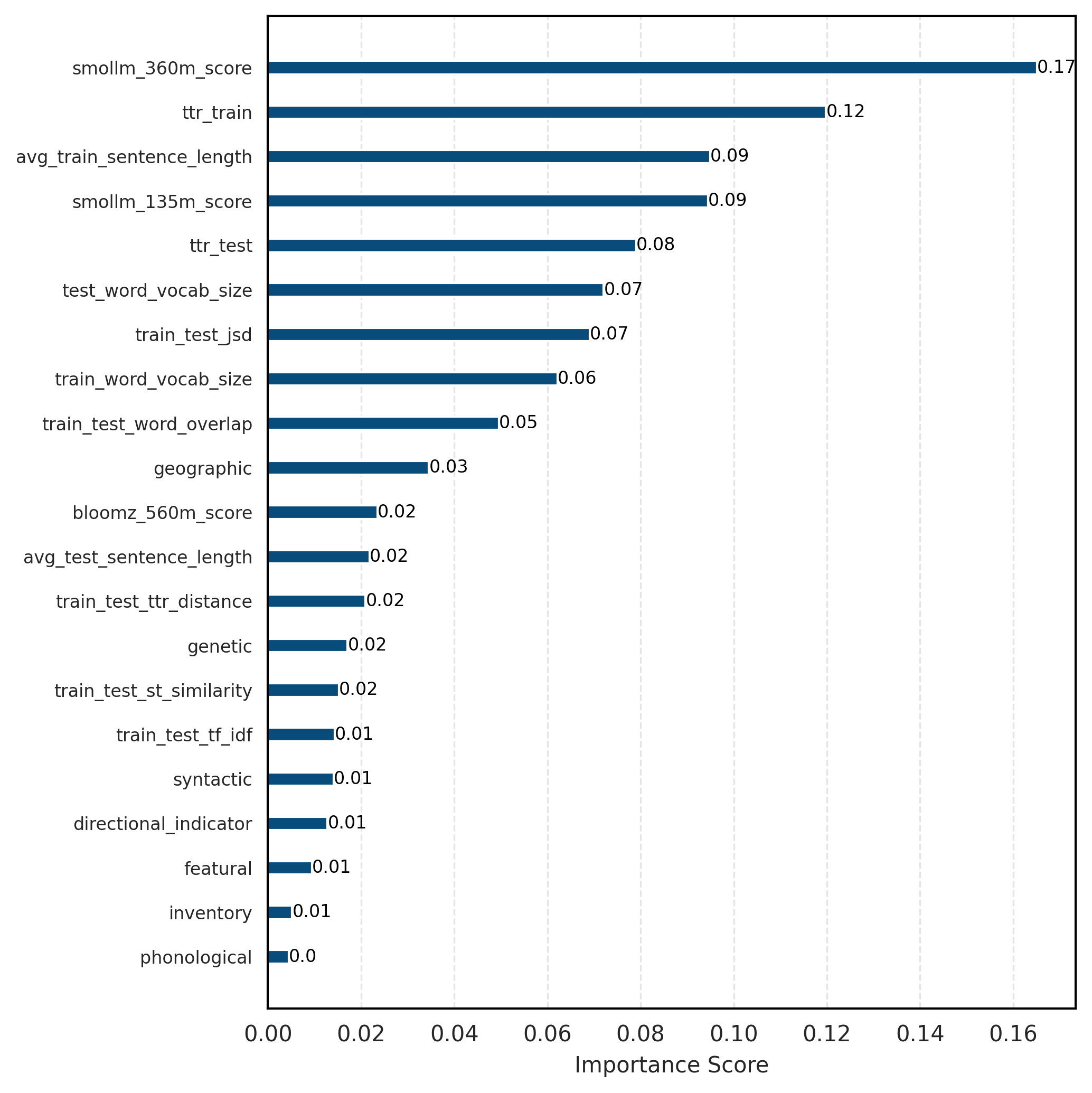}
    \caption{Feature importance analysis of XGBoost with \methodname{} Ensemble on the \textbf{LLaMA3} on \textbf{slot filling} task.}
    \label{fig:slot-llama-feat}
\end{figure*}

\begin{figure*}[!th]
    \centering
    \includegraphics[width=0.9\linewidth]{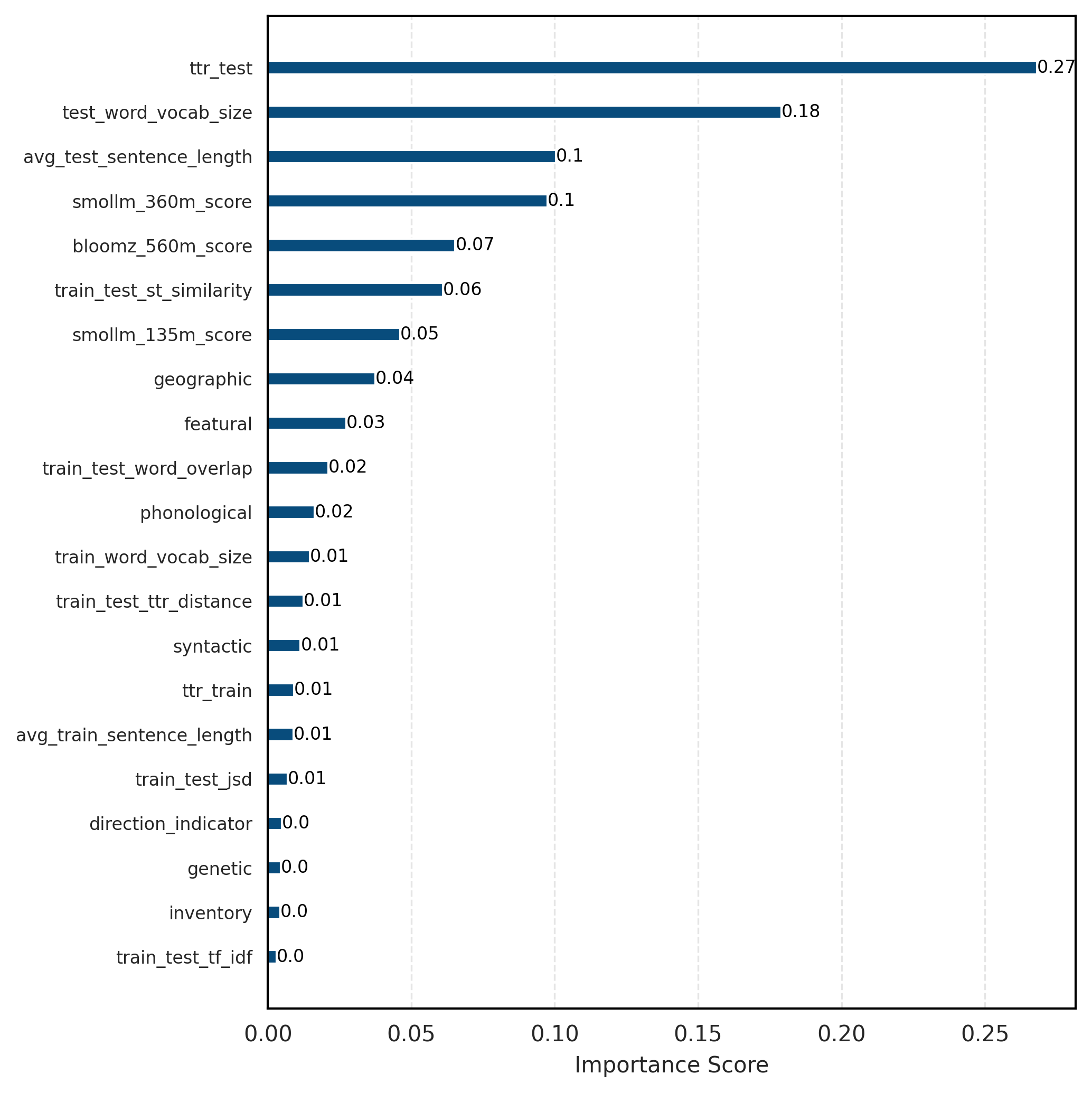}
    \caption{Feature importance analysis of XGBoost with \methodname{} Ensemble on the \textbf{LLaMA3} on \textbf{intent classification} task.}
    \label{fig:intent-llama-feat}
\end{figure*}

We provide feature importance scores of XGBoost with \methodname{} Ensemble for the random English-centric experiment in Figure \ref{fig:m2m100-mt560-feat} and \ref{fig:nllb-mt560-feat}. Each combination consists of one most influential feature followed by others with marginal contributions to the model, each with an importance score of 0.12 or less. We observe that proxy models are always the most influential features in prediction. Other feature importances plot are also provided in Figure \ref{fig:slot-llama-feat} until \ref{fig:intent-llama-feat}.

\section{License for Artifacts}

We discuss the license or terms for the use of any artifacts we use in Table~\ref{tab:datasets-license}.

\begin{table}[!th]
\centering
\resizebox{.49\textwidth}{!}{
    \begin{tabular}{l|cc}
        \toprule
        \textbf{Hyper-parameter} & \textbf{English-centric} & \textbf{Many-to-Many Langs.} \\
        \midrule
        Encoder Layers & 24 & 24 \\
        Decoder Layers & 24 & 24 \\
        Encoder Embed Dim & 1024 & 1024 \\
        Decoder Embed Dim & 1024 & 1024 \\
        Encoder FFN Embed Dim & 8192 & 8192 \\
        Decoder FFN Embed Dim & 8192 & 8192 \\
        Encoder Attention Heads & 16 & 16 \\
        Decoder Attention Heads & 16 & 16 \\
        Encoder Layerdrop & 0.05 & 0.05 \\
        Decoder Layerdrop & 0.05 & 0.05 \\
        Optimizer & Adam & Adam \\
        Adam Eps & 1e-6 & 1e-6 \\
        Adam Betas & (0.9, 0.98) & (0.9, 0.98) \\
        Patience & 6 & 6 \\
        Batch Size & 32 & 32 \\
        Dropout & 0.1 & 0.1 \\
        Attention Dropout & 0.1 & 0.1 \\
        ReLU Dropout & 0.0 & 0.0 \\
        Weight Decay & 0.01 & 0.01 \\
        Label Smoothing & 0.1 & 0.1 \\
        Clip Norm & 1.0 & 1.0 \\
        Learning Rate & 0.00002 & 0.0001 \\
        Max Tokens (per GPU) & 1,000 & 1,000 \\
        \bottomrule
    \end{tabular}
}
\caption{List of hyper-parameters used for NLLB with English-centric and Many-to-Many Languages datasets.}
\label{hyperparam-nllb}
\end{table}

\begin{table}[!th]
\centering
\resizebox{.49\textwidth}{!}{
    \begin{tabular}{l|cc}
        \toprule
        \textbf{Hyper-parameter} & \textbf{English-centric} & \textbf{Many-to-Many Langs.} \\
        \midrule
        Encoder Layers & 6 & 6 \\
        Decoder Layers & 6 & 6 \\
        Encoder Embed Dim & 512 & 512 \\
        Decoder Embed Dim & 512 & 512 \\
        Encoder FFN Embed Dim & 2048 & 2048 \\
        Decoder FFN Embed Dim & 2048 & 2048 \\
        Encoder Attention Heads & 8 & 8 \\
        Decoder Attention Heads & 8 & 8 \\
        Encoder Layerdrop & 0.05 & 0.05 \\
        Decoder Layerdrop & 0.05 & 0.05 \\
        Optimizer & Adam & Adam \\
        Adam Eps & 1e-6 & 1e-6 \\
        Adam Betas & (0.9, 0.98) & (0.9, 0.98) \\
        Patience & 6 & 6 \\
        Batch Size & 32 & 32 \\
        Dropout & 0.1 & 0.1 \\
        Attention Dropout & 0.1 & 0.1 \\
        ReLU Dropout & 0.1 & 0.1 \\
        Weight Decay & 0.0001 & 0.0001 \\
        Label Smoothing & 0.1 & 0.1 \\
        Clip Norm & 0 & 0 \\
        Learning Rate & 0.001 & 0.0005 \\
        Max Tokens (per GPU) & 1,000 & 1,000 \\
        \bottomrule
    \end{tabular}
}
\caption{List of hyper-parameters used for Transformer with English-centric and Many-to-Many Languages datasets.}
\label{hyperparam-transformer}
\end{table}

\begin{table}[!th]
\centering
\resizebox{.49\textwidth}{!}{
    \begin{tabular}{l|cc|cc}
        \toprule
        \textbf{Hyper-parameter} & \multicolumn{2}{c}{\textbf{Using spBLEU}} & \multicolumn{2}{c}{\textbf{Using COMET}}  \\
        & \textbf{M2M100} & \textbf{NLLB} & \textbf{M2M100} & \textbf{NLLB} \\
        \midrule
        max n\_estimators & 5000 & 5000 & 5000 & 5000 \\
        max eta & 0.1 & 0.1 & 0.1 & 0.1 \\
        min\_child\_weight & 5.0 & 4.2 & 3.2 & 1.1 \\
        max\_depth & 5 & 4 & 3 & 5 \\
        gamma & 0 & 0 & 0 & 0 \\
        subsample & 0.6 & 0.94 & 0.6 & 1 \\
        colsample\_bytree & 0.83 & 0.82 & 0.9 & 0.86 \\
        reg\_alpha & 0.2 & 0.32 & 0.11 & 0 \\
        reg\_lambda & 0.1 & 0.37 & 0.48 & 0.05 \\
        \bottomrule
    \end{tabular}
}
\caption{List of hyper-parameters used for XGBoost Regressor on MT task with M2M100 and NLLB models trained with English-centric dataset.}
\label{hyperparam-xgb-mt-eng}
\end{table}

\begin{table}[!th]
\centering
\resizebox{.49\textwidth}{!}{
    \begin{tabular}{l|cc}
        \toprule
        \textbf{Hyper-parameter} & \textbf{M2M100} & \textbf{NLLB}  \\
        \midrule
        max n\_estimators & 2000 & 2000 \\
        max eta & 0.1 & 0.1 \\
        min\_child\_weight & 5 & 2.5 \\
        max\_depth & 3 & 3 \\
        gamma & 0 & 0 \\
        subsample & 0.7 & 0.9 \\
        colsample\_bytree & 0.6 & 0.6 \\
        reg\_alpha & 0 & 0 \\
        reg\_lambda & 0.35 & 0.15 \\
        \bottomrule
    \end{tabular}
}
\caption{List of hyper-parameters used for XGBoost Regressor on MT task with M2M100 and NLLB models trained with Many-to-Many Languages dataset.}
\label{hyperparam-xgb-mt-many}
\end{table}

\begin{table}[!th]
\centering
\resizebox{.49\textwidth}{!}{
    \begin{tabular}{l|cc}
        \toprule
        \textbf{Hyper-parameter} & \textbf{Aya-23} & \textbf{LLaMA3} \\
        \midrule
        max n\_estimators & 5000 & 5000 \\
        max eta & 0.1 & 0.1 \\
        min\_child\_weight & 3.0 & 3.0 \\
        max\_depth & 3 & 3 \\
        gamma & 0.1 & 0.1 \\
        subsample & 0.85 & 0.6 \\
        colsample\_bytree & 1.0 & 0.95 \\
        reg\_alpha & 0.1 & 0.1 \\
        reg\_lambda & 0.2 & 0.5 \\
        \bottomrule
    \end{tabular}
}
\caption{List of hyper-parameters used for XGBoost Regressor on intent classification and slot filling tasks with Aya-23 and LLaMA-3 models.}
\label{hyperparam-xgb-intent-slot}
\end{table}

\begin{table}[!th]
\centering
\resizebox{.3\textwidth}{!}{
    \begin{tabular}{l|c}
        \toprule
        \textbf{Hyper-parameter} & \textbf{Value} \\
        \midrule
        alpha & 0.1 \\
        l1\_ratio & 0.9 \\
        \bottomrule
    \end{tabular}
}
\caption{List of hyper-parameters used for Poly2/Poly3 Regressor for all tasks.}
\label{hyperparam-poly}
\end{table}

\begin{table}[!th]
\centering
\resizebox{.3\textwidth}{!}{
    \begin{tabular}{l|c}
        \toprule
        \textbf{Hyper-parameter} & \textbf{Value} \\
        \midrule
        max learning\_rate & 0.3 \\
        max num\_leaves & 64 \\
        n\_estimators & 100 \\
        max\_bin & 200000 \\
        max\_depth & 10 \\
        min\_child\_weight & 0.001 \\
        min\_child\_samples & 20 \\
        min\_split\_gain & 0.0 \\
        colsample\_bytree & 1.0 \\
        subsample & 1.0 \\
        reg\_alpha & 0.1 \\
        reg\_lambda & 0.1 \\
        \bottomrule
    \end{tabular}
}
\caption{List of hyper-parameters used for LGBM Regressor for all tasks. ``Max" indicates the maximum value set for the hyper-parameter during the hyper-parameter search.}
\label{hyperparam-lgbm}
\end{table}

\begin{table}[!th]
\centering
\fontsize{9}{11}\selectfont
    \begin{tabular}{l|c}
        \toprule
        \textbf{Hyper-parameter} & \textbf{Specification} \\
        \midrule
        max alpha & 0.01 \\
        beta\_w & 0.1 \\
        beta\_h & 0.1 \\
        beta\_z & 0.01 \\
        beta\_s & 0.01 \\
        beta\_t & 0.01 \\
        lr\_decay & 0.001 \\
        iterations & 2000 \\
        \bottomrule
    \end{tabular}
\caption{List of hyper-parameters used for MF Regressor with M2M100 and NLLB models trained with Many-to-Many Languages datasets. ``Max" indicates the maximum value set for the hyper-parameter during the hyper-parameter search.}
\label{hyperparam-mf}
\end{table}

\begin{table}[!th]
\centering
\resizebox{.49\textwidth}{!}{
    \begin{tabular}{lrrrrrr}
    \toprule   
    \textbf{Experimental Settings} & \textbf{Train Size} & \textbf{Test Size} \\
    \midrule
    Random (English-centric) & 1,367 & 587 \\
    Random (Many-to-Many Langs.) & 156 & 68  \\
    Random (Intent Classification) & 1,820 & 781 \\
    Random (Slot Filling) & 1,820 & 781 \\
    Unseen & 1,853 & 101 \\
    Cross-Dataset & 1,954 & 224 \\ \bottomrule
    \end{tabular}
}
\caption{Regressor's training and test set size on different experimental settings. The total MT experimental records for English-centric and Many-to-Many Languages datasets are 1,954 and 224, respectively. On the other hand, the total experimental records for intent classification and slot filling are both 2,601.}
\label{tab:regressor-train-test-size}
\end{table}

\begin{table*}[!ht]
\centering
\resizebox{0.96\textwidth}{!}{
    \begin{tabular}{llc}
    \toprule   
    \textbf{Datasets} & \multicolumn{1}{l}{\textbf{URL Link}} & \textbf{License} \\
    \midrule 
    \multicolumn{1}{l}{MT560~\cite{gowda2021many}} & \url{https://opus.nlpl.eu/MT560} & Unknown \\
    \midrule
    FLoRes~\cite{costa2022no} & \url{Muennighoff/flores200} & CC-BY-SA 4.0 \\ \midrule
    NusaTranslation~\cite{cahyawijaya2023nusawrites} & 
    \url{https://huggingface.co/datasets/indonlp/nusatranslation_mt} & Apache 2.0 \\

    MASSIVE~\cite{fitzgerald2022massive} & \url{https://huggingface.co/datasets/AmazonScience/massive} & CC-BY 4.0
    \\ \bottomrule
    \end{tabular}
}
\caption{List of datasets under study with their licenses.}
\label{tab:datasets-license}
\end{table*}

\begin{figure*}[!th]
    \centering
    \includegraphics[width=1\linewidth]{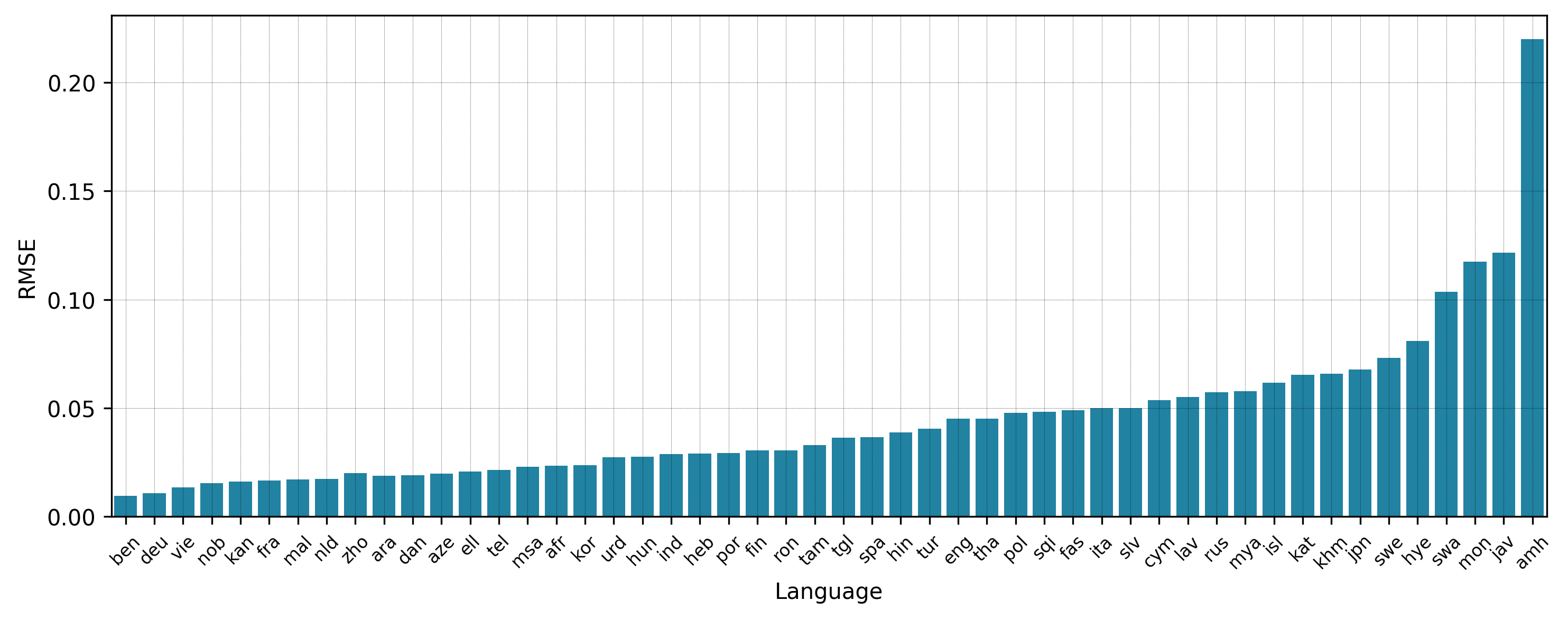}
    \caption{Detailed results of XGBoost with \methodname{} Ensemble on the \textbf{LLaMA3} model under the \textbf{LOLO} setting on \textbf{intent classification} from Table \ref{tab:app-results-intent-detail} per languages.}
    \label{fig:lolo-llama3-intent-by-lang-appendix}
\end{figure*}

\begin{figure*}[!th]
    \centering
    \includegraphics[width=0.9\linewidth]{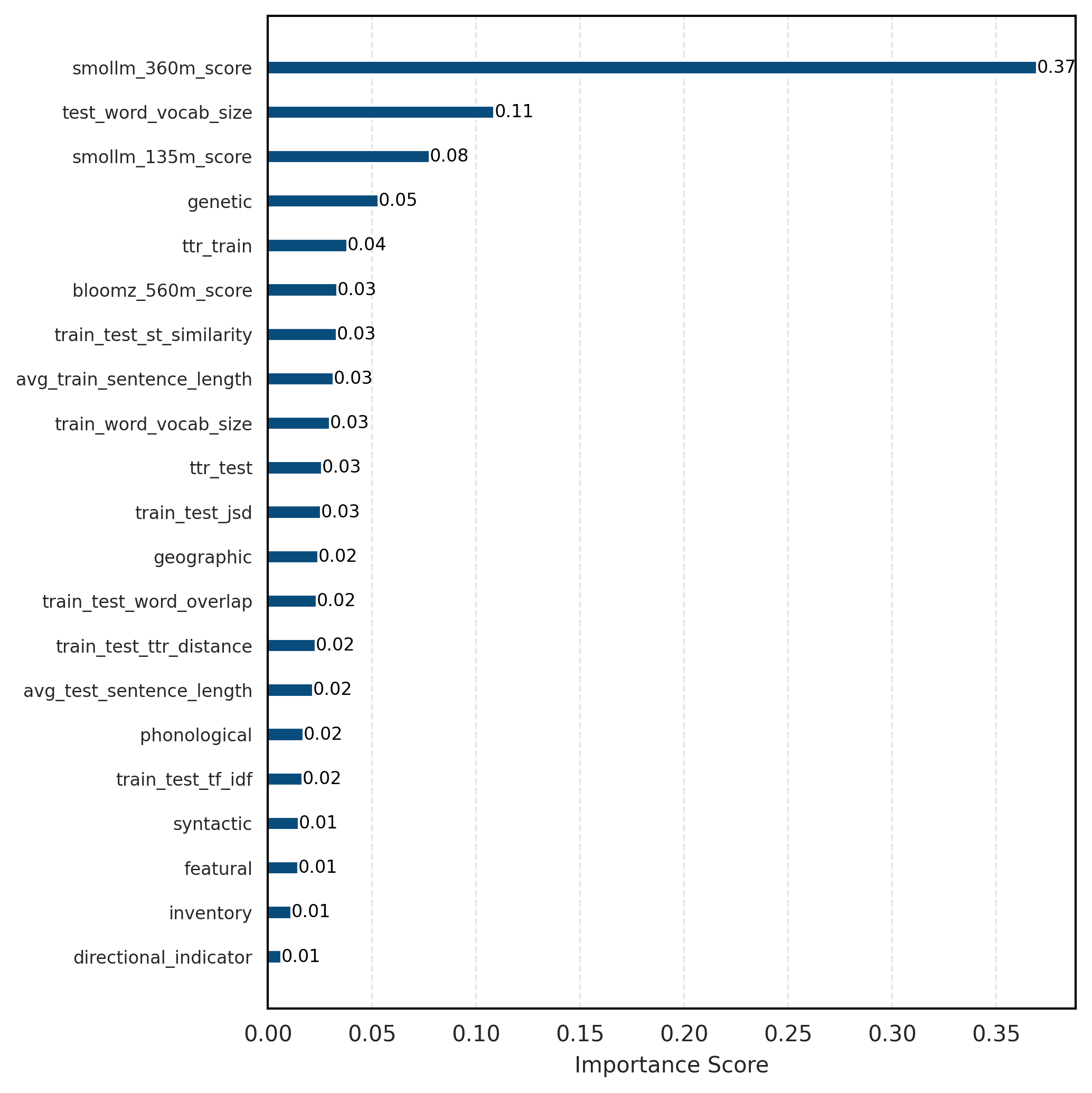}
    \caption{Feature importance analysis of XGBoost with \methodname{} Ensemble on the \textbf{Aya-23} on \textbf{slot filling} task.}
    \label{fig:slot-aya-feat}
\end{figure*}

\begin{figure*}[!th]
    \centering
    \includegraphics[width=0.9\linewidth]{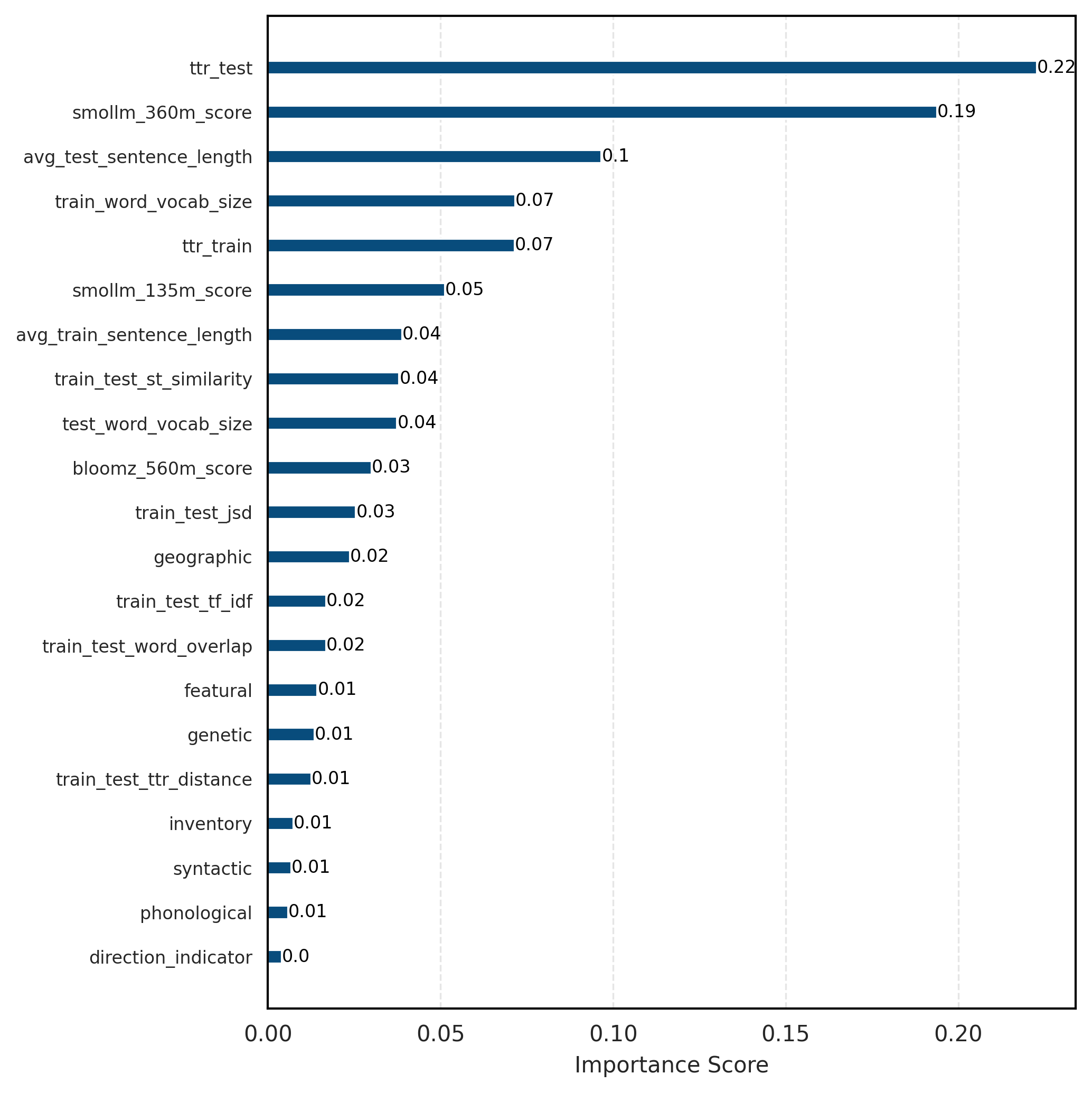}
    \caption{Feature importance analysis of XGBoost with \methodname{} Ensemble on the \textbf{Aya-23} on \textbf{intent classification} task.}
    \label{fig:intent-aya-feat}
\end{figure*}

\end{document}